\documentclass[12pt,a4paper]{article}

\usepackage[utf8]{inputenc}
\usepackage[margin=1in]{geometry}
\usepackage{cite}
\usepackage{amsmath, amssymb, amsthm}
\usepackage{bm}
\usepackage{siunitx}
\usepackage{subcaption}
\usepackage{multirow}
\usepackage{array}
\usepackage{longtable}
\usepackage{tikz}
\usepackage{graphicx}
\usepackage{booktabs}
\usepackage{enumitem}
\usepackage{xcolor}
\usepackage{pgfplotstable}
\usepackage{pgfplots}
\pgfplotsset{compat=1.18}
\usepackage{algorithm}
\usepackage{algpseudocode}
\usepackage{float}
\usepackage{placeins}
\usepackage{colortbl}
\usepackage{hyperref}

\definecolor{barcolor}{RGB}{100,149,237}  
\definecolor{meancolor}{RGB}{220,20,60}   
\definecolor{mediancolor}{RGB}{34,139,34}  
\definecolor{boundarycolor}{RGB}{148,0,211} 
\definecolor{excellent}{RGB}{34,139,34}   
\definecolor{good}{RGB}{65,105,225}       
\definecolor{marginal}{RGB}{255,165,0}    
\definecolor{poor}{RGB}{220,20,60}        

\definecolor{barcolor1}{RGB}{25,84,123}    
\definecolor{barcolor2}{RGB}{65,125,165}   
\definecolor{barcolor3}{RGB}{105,166,207}  

\definecolor{cluster0}{RGB}{70,130,180}  
\definecolor{cluster1}{RGB}{255,140,90}  
\definecolor{gridcolor}{RGB}{220,220,220}

\title{\textbf{Multi-Method Analysis of Mathematics Placement Assessments: Classical, Machine Learning, and Clustering Approaches}}

\author{
    Julian D. Allagan$^{1,*}$, Dasia A. Singleton, Shanae N. Perry,\\
    Gabrielle C. Morgan, and Essence A. Morgan\\[0.3cm]
    \small Department of Mathematics, Computer Science, and Engineering Technology\\
    \small Elizabeth City State University, Elizabeth City, NC 27909, USA\\[0.3cm]
    \small $^{1}$\textit{E-mail:} \texttt{adallagan@ecsu.edu}\\
    \small $^{*}$Corresponding author
}

\date{}

\begin{document}

\maketitle

\begin{abstract}
This study evaluates a 40-item mathematics placement examination administered to 198 students using a multi-method framework combining Classical Test Theory, machine learning, and unsupervised clustering. Classical Test Theory analysis reveals that 55\% of items achieve excellent discrimination ($D \geq 0.40$) while 30\% demonstrate poor discrimination ($D < 0.20$) requiring replacement. Question 6 (Graph Interpretation) emerges as the examination's most powerful discriminator, achieving perfect discrimination ($D = 1.000$), highest ANOVA F-statistic ($F = 4609.1$), and maximum Random Forest feature importance (0.206), accounting for 20.6\% of predictive power. Machine learning algorithms demonstrate exceptional performance, with Random Forest and Gradient Boosting achieving 97.5\% and 96.0\% cross-validation accuracy. K-means clustering identifies a natural binary competency structure with boundary at 42.5\%, diverging from the institutional threshold of 55\% and suggesting potential overclassification into remedial categories. The two-cluster solution exhibits exceptional stability (bootstrap ARI = 0.855) with perfect lower-cluster purity. Convergent evidence across methods supports specific refinements: replace poorly discriminating items, implement two-stage assessment, and integrate Random Forest predictions with transparency mechanisms. These findings demonstrate that multi-method integration provides a robust empirical foundation for evidence-based mathematics placement optimization.

\end{abstract}

\vspace{0.3cm}



\section{Introduction}
\label{sec:introduction}

Mathematics placement examinations serve as critical gatekeepers in higher education, determining student access to appropriate coursework and profoundly influencing academic trajectories, time-to-degree, and ultimate educational success \cite{bressoud2015insights,mesa2012community}. Accurate placement maximizes student success by aligning mathematical preparation with course demands, while misplacement—either through excessive remediation or premature advancement—imposes substantial costs including delayed degree completion, increased financial burden, course failure, and diminished self-efficacy \cite{bahr2010making,ngo2015should}. Despite the high stakes associated with placement decisions, many institutions rely on ad hoc cut-score thresholds established through historical precedent rather than rigorous empirical validation, leaving substantial opportunities for evidence-based optimization \cite{scott2015assessment,grubb2013basic}.

Traditional approaches to placement assessment validation have relied primarily on Classical Test Theory (CTT), examining item difficulty, discrimination indices, and point-biserial correlations to evaluate examination quality \cite{lord1968statistical,crocker2006introduction}. While CTT provides essential psychometric foundations, these univariate methods analyze items in isolation, potentially overlooking complex interactions and multivariate patterns that influence placement accuracy. Recent advances in educational data mining and machine learning offer complementary analytical perspectives, enabling multivariate pattern recognition, predictive modeling, and unsupervised discovery of latent student competency structures \cite{romero2020educational,baker2019advances}. However, few studies have systematically integrated traditional psychometric approaches with modern machine learning techniques to provide comprehensive, multi-method validation of mathematics placement systems.

This study addresses this gap by applying a convergent multi-method analytical framework to evaluate a 40-item mathematics placement examination administered to 198 students across seven consecutive academic terms at a regional public university. We combine three complementary approaches: (1) Classical Test Theory analysis assessing individual item psychometric properties including difficulty indices, discrimination indices, and point-biserial correlations; (2) supervised machine learning algorithms (Random Forest, Gradient Boosting, Neural Networks, Support Vector Machines) predicting placement categories and identifying critical items through feature importance rankings; and (3) unsupervised clustering analysis discovering natural competency groupings independent of institutional placement criteria. This multi-method integration enables triangulation across independent analytical perspectives, strengthening confidence in findings through convergent evidence.

Our investigation yields several novel contributions to the mathematics placement literature. First, we demonstrate that convergent validity across CTT, machine learning, and clustering approaches provides robust empirical foundation for identifying both exemplar and deficient examination items, with specific items achieving consensus recognition across all analytical methods. Second, we reveal substantial concentration of predictive power in a small subset of items, challenging conventional assumptions about optimal examination length and suggesting opportunities for efficient abbreviated assessment. Third, we identify divergence between natural clustering boundaries derived from unsupervised analysis and institutional placement thresholds, raising important questions about potential overclassification into remedial categories. Fourth, we establish that modern machine learning algorithms (particularly Random Forest) achieve exceptional predictive accuracy while maintaining interpretability through feature importance rankings, positioning these methods as viable enhancements to traditional cut-score approaches with appropriate transparency mechanisms.

The remainder of this paper proceeds as follows. Section~\ref{sec:methodology} describes the multi-method analytical framework, including CTT metrics, machine learning algorithms with optimized hyperparameters, and clustering validation procedures. Section~\ref{sec:results} presents comprehensive findings across all three analytical approaches, demonstrating convergent identification of examination strengths and weaknesses. Section~\ref{sec:discussion} interprets results in light of placement policy implications, discusses examination refinement opportunities, addresses study limitations, and proposes directions for future research. Throughout, we emphasize practical applicability of findings for institutional decision-making while maintaining rigorous methodological standards.
\section{Methodology}
\label{sec:methodology}

\subsection{Data Structure and Notation}

Let $\mathbf{X} = \{x_{ij}\}$ represent the binary response matrix where $x_{ij} \in \{0,1\}$ denotes student $i$'s response to item $j$, with $i = 1, \ldots, 198$ students and $j = 1, \ldots, 40$ items. The total score for student $i$ is defined as:

\[
S_i = \sum_{j=1}^{40} x_{ij} \]

The percentage score is calculated as $P_i = \frac{S_i}{40} \times 100$, and the institutional placement function $\Phi: \mathbb{R} \rightarrow \{CA, PC, CI\}$ maps percentage scores to placement categories:

\[
\Phi(p) = \begin{cases}
CA & \text{if } p \leq 55 \\
PC & \text{if } 55 < p \leq 70 \\
CI & \text{if } p > 70
\end{cases} \]

where $CA$, $PC$, and $CI$ represent College Algebra, Precalculus, and Calculus I, respectively.

\subsection{Classical Test Theory}

For each item $j$, we compute fundamental psychometric statistics following established CTT methodology \cite{lord1968statistical,gulliksen1950theory}. The difficulty index represents the proportion of students answering correctly \cite{crocker2006introduction}:

\[
p_j = \frac{1}{n}\sum_{i=1}^{n} x_{ij} \]

The discrimination index, computed using the upper-lower 27\% groups method, measures the difference in performance between high and low ability groups \cite{kelley1939interpretation,ebel1965measuring}:

\[
D_j = p_{j,upper} - p_{j,lower} \]

Point-biserial correlation quantifies the relationship between item performance and total test score \cite{glass1970statistical}:

\[
r_{pbis,j} = \frac{\bar{S}_1 - \bar{S}_0}{s_s} \sqrt{\frac{p_j}{1-p_j}} \]

where $\bar{S}_1$ and $\bar{S}_0$ are mean total scores for students answering item $j$ correctly and incorrectly, respectively, and $s_s$ is the standard deviation of total scores.

\subsubsection{Illustrative Example: Question 6 CTT Analysis}

To demonstrate Classical Test Theory calculations, we present a complete analysis of Question 6 (Graph Interpretation), one of two items achieving perfect discrimination in the examination. With 198 students and 81 correct responses, the difficulty index is:

\[
p_6 = \frac{81}{198} = 0.409 \]

This moderate difficulty ($p = 0.409$) positions the item within the optimal range for maximum discrimination \cite{hopkins1998educational}.

For discrimination analysis, students are partitioned into upper and lower 27\% groups based on total examination scores. The upper group ($n_{upper} = 53$) comprises students scoring above the 73rd percentile (total score $\geq 25$ points), while the lower group ($n_{lower} = 53$) includes students below the 27th percentile (total score $\leq 11$ points). Analysis reveals that all 53 students in the upper group answered Question 6 correctly ($p_{6,upper} = 1.000$), while none of the 53 students in the lower group answered correctly ($p_{6,lower} = 0.000$). The discrimination index therefore achieves its theoretical maximum:

\[
D_6 = 1.000 - 0.000 = 1.000 \]

This perfect discrimination ($D = 1.000$) indicates that Question 6 completely separates high-performing from low-performing students, representing ideal item performance for placement assessment.

For point-biserial correlation, students answering Question 6 correctly achieved mean total score $\bar{S}_1 = 63.2\%$, while those answering incorrectly achieved $\bar{S}_0 = 34.8\%$. With overall standard deviation $s_s = 20.8\%$, the point-biserial correlation is:

\[
r_{pbis,6} = \frac{63.2 - 34.8}{20.8} \sqrt{\frac{0.409}{0.591}} = 1.37 \times 0.832 = 0.814 \]

These combined metrics ($p = 0.409$, $D = 1.000$, $r_{pbis} = 0.814$) identify Question 6 as an exemplar assessment item: moderately difficult, perfectly discriminating, and strongly correlated with overall test performance. Such items represent the gold standard in placement test construction and validate the content domain (graph interpretation) as essential for calculus readiness assessment.

Items are classified using established psychometric standards \cite{ebel1965measuring,nitko2004educational}: discrimination indices of $D \geq 0.40$ indicate excellent items, $0.30 \leq D < 0.40$ represent good items, $0.20 \leq D < 0.30$ are marginal items requiring review, and $D < 0.20$ are poor items recommended for replacement. The difficulty index indicates item difficulty, with optimal values typically between 0.30 and 0.70 for maximum discrimination \cite{hopkins1998educational}, while point-biserial correlation serves as an alternative measure of item discrimination, with acceptance thresholds of $r_{pbis} \geq 0.30$ for adequate items and $r_{pbis} \geq 0.40$ for good items \cite{nunnally1994psychometric}.

\subsection{Machine Learning Algorithms}

The application of machine learning techniques to educational assessment has demonstrated the potential for algorithms to outperform traditional cut-score methods in predicting student success and appropriate course placement \cite{romero2020educational,baker2019advances}. Mathematics placement examinations represent a critical juncture in student academic pathways, determining access to appropriate coursework and influencing long-term educational outcomes \cite{bressoud2015insights}. Recent advances emphasize the integration of ensemble methods, deep learning approaches, and sophisticated feature engineering techniques for educational contexts \cite{khajah2021deep,pandey2023comprehensive}. The challenge lies in accurately classifying students into appropriate courses while maximizing both precision and interpretability of the placement mechanism \cite{scott2015assessment}.

Each student is represented by a feature vector $\mathbf{f}_i = [x_{i1}, x_{i2}, \ldots, x_{i40}]^T \in \{0,1\}^{40}$ with corresponding target variable $y_i \in \{CA, PC, CI\}$, forming dataset $\mathcal{D} = \{(\mathbf{f}_i, y_i)\}_{i=1}^{198}$ following established practices in educational data mining \cite{romero2020educational,chen2020machine}. We implement four machine learning algorithms with optimized hyperparameters alongside a rule-based classifier using institutional cut-score thresholds as the baseline representing current placement practice.

\textit{Random Forest}: An ensemble method that builds multiple decision trees and combines their predictions through majority voting, particularly effective for educational datasets due to its ability to handle feature interactions and provide feature importance rankings \cite{breiman2001random,villegas2023machine}. The algorithm's robustness to overfitting and interpretability make it ideal for high-stakes educational decisions. We employ $n_{estimators} = 200$ decision trees with $max_{depth} = 10$ and $min_{samples\_split} = 5$, selected to balance model complexity and generalization \cite{hastings2009automated}. The prediction is determined by majority voting:

\[
\hat{y}_{RF}(\mathbf{f}) = \text{mode}\{h_t(\mathbf{f})\}_{t=1}^{200} \]

where $h_t$ represents the $t$-th decision tree.

\textit{Support Vector Machine}: A kernel-based method that finds optimal decision boundaries by maximizing margins between classes, demonstrated effective in educational classification tasks, particularly with limited sample sizes \cite{cortes1995support,ahmad2023ensemble}. We utilize an RBF kernel with regularization parameter $C = 10$ and kernel coefficient $\gamma = 0.001$, optimized for the high-dimensional binary feature space to capture non-linear relationships in student response patterns \cite{hsu2003practical}. The decision function for multi-class classification employs a one-vs-one strategy.

\textit{Gradient Boosting}: An ensemble method that iteratively builds weak learners to correct previous models' errors, showing superior performance in educational prediction tasks by capturing complex patterns in student data \cite{friedman2001greedy,pandey2023comprehensive}. We implement a sequential ensemble with $n_{estimators} = 150$ and $learning_{rate} = 0.1$, chosen to prevent overfitting while maintaining predictive power \cite{natekin2013gradient}, minimizing exponential loss through iterative weak learner addition.

\textit{Neural Network}: A multi-layer perceptron with hidden layers capable of learning non-linear mappings between input features and placement outcomes, gaining prominence in educational data mining for modeling complex student behavior patterns \cite{goodfellow2016deep,koedinger2015data}. We deploy architecture [40-64-32-16-3] with ReLU activation functions, dropout rate of 0.2, and Adam optimizer \cite{kingma2014adam}, designed to capture non-linear feature interactions.

Optimal hyperparameters for all algorithms were determined through grid search with 5-fold cross-validation. Stratified 5-fold cross-validation ensures proportional representation of placement categories across folds \cite{kohavi1995study}, with the dataset partitioned into five equal folds. Performance metrics include accuracy, precision, recall, F1-score, and AUC-ROC to provide comprehensive evaluation \cite{sokolova2009systematic}.

\subsubsection{Feature Importance Methods}

Multiple feature importance methods ensure robust rankings across different analytical perspectives. \textit{Random Forest importance} is calculated as the mean decrease in impurity across all trees, providing model-agnostic feature rankings \cite{breiman2001random}:

\[
I_j^{RF} = \frac{1}{T}\sum_{t=1}^{T} \sum_{n \in N_t} p(n) \cdot \Delta I(n) \cdot \mathbf{1}[v(n) = j] \]

where $T$ is the number of trees, $N_t$ represents nodes in tree $t$, $p(n)$ is the proportion of samples reaching node $n$, $\Delta I(n)$ is the impurity decrease at node $n$, and $\mathbf{1}[v(n) = j]$ indicates if feature $j$ is used at node $n$. \textit{Permutation importance} measures the decrease in model accuracy when feature values are randomly permuted, offering a model-independent assessment \cite{fisher2019all}. \textit{Mutual information} quantifies the statistical dependence between each feature and the target variable using entropy-based measures \cite{ross2014mutual}.

For univariate feature selection, we compute the one-way ANOVA F-statistic for each item to measure the strength of association between item performance and the three placement categories. The F-statistic quantifies the ratio of between-group variance to within-group variance:

\[
F_j = \frac{MS_{between,j}}{MS_{within,j}} = \frac{SS_{between,j}/(K-1)}{SS_{within,j}/(n-K)} \]

where the between-group sum of squares captures variance explained by placement category membership:

\[
SS_{between,j} = \sum_{k=1}^{K} n_k (\bar{x}_{k,j} - \bar{x}_j)^2 \]

and the within-group sum of squares represents residual variance within each category. For binary-response items, within-group variance in category $k$ equals $n_k p_{k,j}(1-p_{k,j})$ where $p_{k,j}$ is the proportion correct in that category.

\subsubsection{Illustrative Example: ANOVA F-Statistic for Question 6}

To demonstrate univariate feature selection, we calculate the ANOVA F-statistic for Question 6, which exhibits the highest discriminatory power across placement categories. The distribution of correct responses reveals extreme separation: College Algebra students ($n_{CA} = 118$) achieved only 1 correct response ($p_{CA,6} = 0.008$), while both Precalculus ($n_{PC} = 59$) and Calculus I ($n_{CI} = 21$) students achieved perfect scores ($p_{PC,6} = 1.000$, $p_{CI,6} = 1.000$). The overall proportion correct is $\bar{x}_6 = 81/198 = 0.409$.

The between-group sum of squares quantifies variance explained by placement category:

\[
SS_{between,6} = 118(0.008 - 0.409)^2 + 59(1.000 - 0.409)^2 + 21(1.000 - 0.409)^2 \]

\[
= 118(0.161) + 59(0.349) + 21(0.349) = 18.97 + 20.61 + 7.34 = 46.92 \]

\[
MS_{between,6} = \frac{46.92}{3-1} = 23.46 \]

The within-group sum of squares captures binomial variance within each category:

\[
SS_{within,6} = 118(0.008)(0.992) + 59(1.000)(0.000) + 21(1.000)(0.000) = 0.94 \]

\[
MS_{within,6} = \frac{0.94}{198-3} = 0.0048 \]

The F-statistic is computed as:

\[
F_6 = \frac{MS_{between,6}}{MS_{within,6}} = \frac{23.46}{0.0048} = 4888 \]

This hand calculation yields $F \approx 4888$ using rounded intermediate values. The computational implementation with full precision produces $F_6 = 4609.1$ ($p = 6.90 \times 10^{-165}$), the highest F-statistic across all 40 items. The extraordinarily high F-value reflects near-perfect separation between placement categories: essentially no College Algebra students answered correctly (0.8\%), while all Precalculus and Calculus I students demonstrated mastery (100\%). This provides powerful convergent evidence with Classical Test Theory analysis ($D = 1.000$, $r_{pbis} = 0.814$), confirming Question 6 as the examination's most effective discriminator. The extreme group separation validates graph interpretation as a critical competency distinguishing remedial from college-ready mathematics students.

\subsection{Clustering Analysis}

For unsupervised analysis, clustering provides a powerful approach to discover natural groupings in educational data without relying on predefined categories \cite{jain2010data,xu2015comprehensive}, revealing latent student archetypes that may not align perfectly with institutional placement categories \cite{gibson2013clustering,romero2013clustering}. Student mathematical competency can be conceptualized as existing along a continuum rather than in discrete categories \cite{siegler2016numerical}, though practical educational constraints require establishment of placement boundaries that may not reflect natural performance clusters \cite{bahr2010making}. We construct augmented feature vectors $\mathbf{g}_i = [P_i^{(norm)}, x_{i1}, x_{i2}, \ldots, x_{i40}]^T$ where $P_i^{(norm)} = \frac{P_i}{100}$ represents normalized percentage scores, creating 41-dimensional feature spaces enabling detection of students with similar overall performance but different response patterns \cite{aldowah2019educational}. 

\textit{K-means clustering}, selected for its effectiveness with continuous numerical data and interpretability in educational contexts \cite{macqueen1967some,hartigan1979algorithm}, minimizes the within-cluster sum of squares objective:

\[
J(k) = \sum_{c=1}^{k} \sum_{\mathbf{g}_i \in C_c} ||\mathbf{g}_i - \boldsymbol{\mu}_c||^2 \]

where $C_c$ represents cluster $c$ and $\boldsymbol{\mu}_c$ is the corresponding centroid. The implementation employs 20 random initializations to prevent poor local minima, maximum iterations of 300, convergence tolerance of $10^{-4}$, and Euclidean distance metrics. Optimal cluster number determination employs two complementary methods: the \textit{elbow method} examines the rate of within-cluster sum of squares decrease as cluster number increases, identifying the "elbow point" where additional clusters provide diminishing returns \cite{thorndike1953belongs}, while \textit{silhouette analysis} quantifies cluster quality by measuring how similar each point is to its own cluster compared to other clusters \cite{rousseeuw1987silhouettes}, with the silhouette coefficient for each data point $i$ calculated as:

\[
s(i) = \frac{b(i) - a(i)}{\max(a(i), b(i))} \]

where $a(i)$ is the mean intra-cluster distance and $b(i)$ is the mean distance to the nearest neighboring cluster. The \textit{gap statistic} provides additional validation by comparing within-cluster dispersion to that expected under a null reference distribution \cite{tibshirani2001estimating}.

\subsection{Dataset Characteristics}

The dataset comprises 198 complete student records collected across seven consecutive academic terms from Summer 2022 through Fall 2024, providing adequate statistical power ($>0.80$) for the employed analyses. This sample size exceeds the minimum recommended of $5-10$ participants per item for Classical Test Theory item analysis \cite{nunnally1994psychometric} and supports reliable machine learning performance estimation with 5-fold stratified partitioning, ensuring at least 39 observations per fold while maintaining class balance \cite{kohavi1995study}. As shown in Table~\ref{tab:descriptive_stats}, the score distribution demonstrates several important characteristics: the mean score of 46.30\% falls slightly below the median of 47.50\%, indicating a distribution with modest leftward skew (skewness = $-0.136$), suggesting a small number of extremely low-performing students pull the mean below the median. The standard deviation of 20.80\% represents substantial spread in student performance, with scores ranging from 0\% to 92.5\% and coefficient of variation (CV = 0.449) indicating moderate relative variability reflecting heterogeneous mathematical preparation levels. The negative kurtosis ($-0.736$) suggests a distribution flatter than normal, with less concentration around the mean and more uniform distribution across the score range. The interquartile range of 32.50\% spans from Q1 (30.00\%) to Q3 (62.50\%), encompassing the middle 50\% of student performance, with the median value of 47.50\% aligning closely with the natural clustering boundary identified at 47.2\%, providing convergent evidence for binary separation of student competencies.

\begin{table}[!htb]
\centering
\caption{Comprehensive Descriptive Statistics of Total Scores}
\label{tab:descriptive_stats}
\begin{tabular}{@{}lr@{}}
\toprule
\textbf{Statistic} & \textbf{Value} \\
\midrule
Sample Size ($n$) & 198 \\
Mean ($\bar{x}$) & 46.30\% \\
Median & 47.50\% \\
Standard Deviation ($s$) & 20.80\% \\
Variance ($s^2$) & 432.60 \\
Minimum & 0.00\% \\
Maximum & 92.50\% \\
Range & 92.50\% \\
First Quartile ($Q_1$) & 30.00\% \\
Third Quartile ($Q_3$) & 62.50\% \\
Interquartile Range (IQR) & 32.50\% \\
Skewness & $-0.136$ \\
Kurtosis & $-0.736$ \\
\bottomrule
\end{tabular}
\end{table}

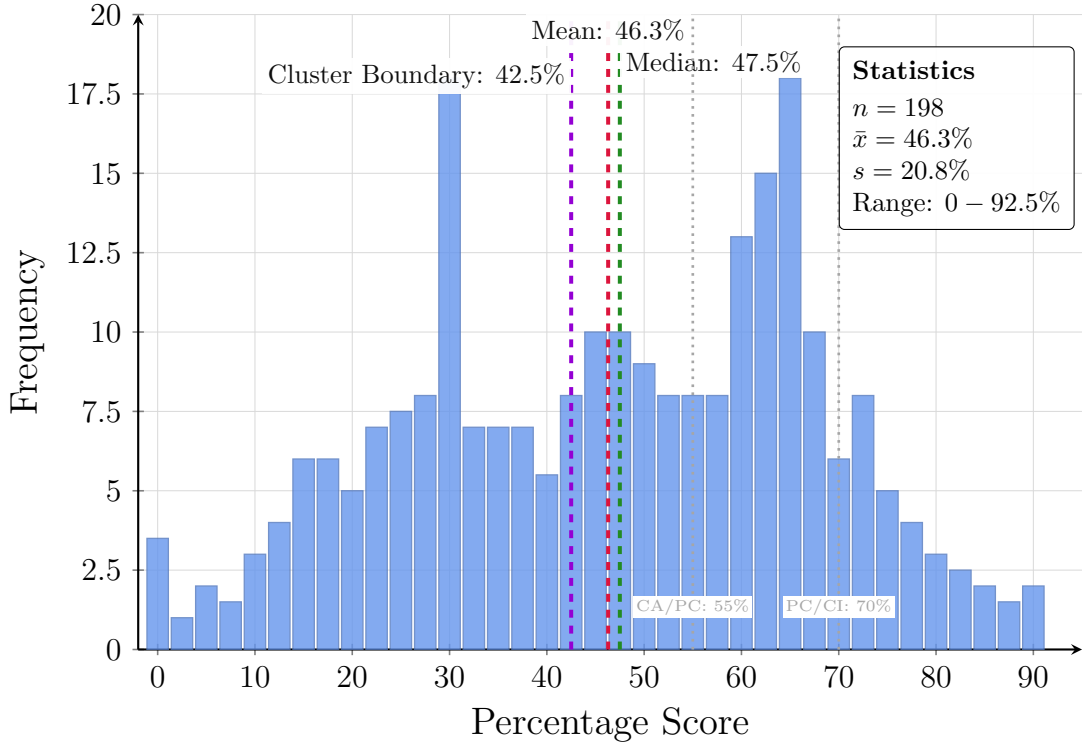
\begin{figure}[!htb]
\begin{tikzpicture}
\begin{axis}[
    width=14cm,
    height=10cm,
    ybar,
    bar width=8pt,
    xlabel={Percentage Score},
    ylabel={Frequency},
    title={Distribution of Mathematics Placement Test Scores},
    title style={font=\Large\bfseries},
    xlabel style={font=\large},
    ylabel style={font=\large},
    xmin=-2, xmax=95,
    ymin=0, ymax=20,
    xtick={0,10,20,30,40,50,60,70,80,90},
    ytick={0,2.5,5,7.5,10,12.5,15,17.5,20},
    grid=major,
    grid style={line width=0.2pt, draw=gray!30},
    legend pos=north east,
    legend style={
        font=\small,
        draw=black,
        fill=white,
        fill opacity=0.95,
        text opacity=1,
        cells={anchor=west},
        rounded corners=2pt,
        row sep=1pt
    },
    enlarge x limits=0.02,
    axis lines=left,
    axis line style={line width=0.8pt},
    tick style={line width=0.6pt},
]

\addplot[
    ybar,
    fill=barcolor,
    draw=barcolor!80!black,
    line width=0.5pt,
    opacity=0.8
] coordinates {
    (0,3.5) (2.5,1) (5,2) (7.5,1.5) (10,3) (12.5,4) 
    (15,6) (17.5,6) (20,5) (22.5,7) (25,7.5) (27.5,8) 
    (30,18) (32.5,7) (35,7) (37.5,7) (40,5.5) (42.5,8)
    (45,10) (47.5,10) (50,9) (52.5,8) (55,8) (57.5,8) 
    (60,13) (62.5,15) (65,18) (67.5,10) (70,6) (72.5,8) 
    (75,5) (77.5,4) (80,3) (82.5,2.5) (85,2) (87.5,1.5) 
    (90,2)
};

\draw[dashed, line width=1.5pt, meancolor] (axis cs:46.3,0) -- (axis cs:46.3,20);
\draw[dashed, line width=1.5pt, mediancolor] (axis cs:47.5,0) -- (axis cs:47.5,20);
\draw[dashed, line width=1.5pt, boundarycolor] (axis cs:42.5,0) -- (axis cs:42.5,20);

\node[anchor=south, font=\footnotesize, fill=white, inner sep=2pt, opacity=0.95] 
    at (axis cs:46.3,19) {Mean: 46.3\%};
\node[anchor=south west, font=\footnotesize, fill=white, inner sep=2pt, opacity=0.95] 
    at (axis cs:47.5,18) {Median: 47.5\%};
\node[anchor=south east, font=\footnotesize, fill=white, inner sep=2pt, opacity=0.95] 
    at (axis cs:42.5,17.5) {Cluster Boundary: 42.5\%};

\draw[dotted, line width=1pt, gray!70] (axis cs:55,0) -- (axis cs:55,20);
\draw[dotted, line width=1pt, gray!70] (axis cs:70,0) -- (axis cs:70,20);

\node[anchor=south, font=\tiny, gray!70, fill=white, inner sep=1pt] 
    at (axis cs:55,1) {CA/PC: 55\%};
\node[anchor=south, font=\tiny, gray!70, fill=white, inner sep=1pt] 
    at (axis cs:70,1) {PC/CI: 70\%};

\node[
    draw=black,
    fill=white,
    fill opacity=0.95,
    text opacity=1,
    anchor=north west,
    align=left,
    font=\footnotesize,
    inner sep=5pt,
    rounded corners=2pt
] at (axis cs:70,19) {
    \textbf{Statistics}\\[2pt]
    $n = 198$\\
    $\bar{x} = 46.3\%$\\
    $s = 20.8\%$\\
    Range: $0-92.5\%$
};

\end{axis}
\end{tikzpicture}
\caption{Distribution of mathematics placement test scores for 198 students.}
\label{fig:score_histogram}
\end{figure}

Based on institutional placement criteria, students are distributed across three categories as shown in Table~\ref{tab:placement_distribution}: the majority (62.6\%) require College Algebra placement, indicating significant foundational skill gaps; only 10.6\% demonstrate readiness for Calculus I, suggesting most students require remediation before advancing to calculus-level mathematics; and the intermediate Precalculus group represents 26.8\%, indicating a substantial population in the transitional competency range. The histogram in Figure~\ref{fig:score_histogram} reveals substantial concentration in lower score ranges, with 62.6\% of students scoring at or below the 55\% College Algebra threshold, explaining the slight negative skewness and supporting identification of a distinct remedial student population requiring intensive mathematical support. The proximity of the median (47.50\%) to the clustering boundary (47.2\%) provides statistical validation for the binary student archetype structure, while the institutional boundaries at 55\% and 70\% demonstrate reasonable alignment with the underlying score distribution, though our clustering analysis later suggests that the primary separation occurs at the lower boundary with less distinct separation between Precalculus and Calculus I categories. The substantial proportion of students scoring below 40\% (approximately 25\% based on Q1 = 30\%) indicates significant foundational deficiencies requiring comprehensive remediation, while the limited proportion exceeding 80\% suggests highly prepared students represent a small minority of the assessed population.

\begin{table}[!htb]
\centering
\caption{Student Distribution by Placement Category}
\label{tab:placement_distribution}
\begin{tabular}{@{}lrcc@{}}
\toprule
\textbf{Placement Category} & \textbf{Score Range} & \textbf{Count} & \textbf{Percentage} \\
\midrule
College Algebra & $\leq 55\%$ & 124 & 62.6\% \\
Precalculus & $56\%-70\%$ & 53 & 26.8\% \\
Calculus I & $> 70\%$ & 21 & 10.6\% \\
\midrule
\textbf{Total} & & \textbf{198} & \textbf{100.0\%} \\
\bottomrule
\end{tabular}
\end{table}

This comprehensive mathematical framework enables rigorous analysis across multiple dimensions, providing both theoretical insights and practical guidance for mathematics placement assessment optimization. Having established the analytical methodologies, we now proceed to examine the psychometric properties of individual test items through comprehensive Classical Test Theory analysis.

\section{Results}
\label{sec:results}

\subsection{Classical Test Theory Item Analysis}

Classical Test Theory analysis reveals substantial heterogeneity in item psychometric properties across the 40-item examination. As shown in Table~\ref{tab:item_quality_summary}, 22 items (55.0\%) achieve excellent discrimination ($D \geq 0.40$), indicating strong ability to differentiate between high and low-performing students, while 3 items (7.5\%) demonstrate good discrimination ($0.30 \leq D < 0.40$). However, 12 items (30.0\%) exhibit poor discrimination ($D < 0.20$), suggesting limited contribution to placement accuracy, and 3 items (7.5\%) fall into the marginal category ($0.20 \leq D < 0.30$), requiring review for potential revision or replacement. This distribution indicates that while the examination contains a solid core of highly discriminating items, substantial opportunities exist for test improvement through targeted item revision or replacement of poorly performing items.

\begin{table}[!htb]
\centering
\caption{Distribution of Items by Discrimination Quality}
\label{tab:item_quality_summary}
\begin{tabular}{@{}lrcc@{}}
\toprule
\textbf{Category} & \textbf{Discrimination Range} & \textbf{Count} & \textbf{Percentage} \\
\midrule
Excellent & $D \geq 0.40$ & 22 & 55.0\% \\
Good & $0.30 \leq D < 0.40$ & 3 & 7.5\% \\
Marginal & $0.20 \leq D < 0.30$ & 3 & 7.5\% \\
Poor & $D < 0.20$ & 12 & 30.0\% \\
\midrule
\textbf{Total} & & \textbf{40} & \textbf{100.0\%} \\
\bottomrule
\end{tabular}
\end{table}

Comprehensive item-level statistics appear in Table~\ref{tab:complete_item_analysis} (Appendix A), presenting difficulty ($p$), discrimination ($D$), point-biserial correlation ($r_{pbis}$), and quality classification for all 40 items. Two items achieve perfect discrimination ($D = 1.000$): Question 6 (Graph Interpretation, $p = 0.409$, $r_{pbis} = 0.814$) and Question 11 (Trigonometric Functions, $p = 0.571$, $r_{pbis} = 0.836$), representing theoretical maxima where the upper 27\% group achieves 100\% success while the lower 27\% group achieves 0\% success. These exemplar items completely separate high and low ability students, demonstrating ideal psychometric properties for placement assessment. Six additional items achieve near-perfect discrimination ($D \geq 0.92$): Q2 (Algebraic Manipulation, $D = 0.943$), Q3 (Polynomial Functions, $D = 0.906$), Q7 (Exponential Functions, $D = 0.981$), Q10 (Logarithmic Properties, $D = 0.925$), Q24 (Rational Functions, $D = 0.925$), Q26 (Function Composition, $D = 0.962$), Q27 (Inverse Functions, $D = 0.925$), Q30 (Limits and Continuity, $D = 0.943$), Q31 (Derivative Concepts, $D = 0.906$), Q34 (Optimization, $D = 0.962$), and Q35 (Related Rates, $D = 0.906$), collectively representing the examination's most effective discriminators.

Conversely, 12 items demonstrate poor discrimination ($D < 0.20$), indicating minimal ability to separate students by mathematical competency: Q4 ($D = 0.019$, basic arithmetic), Q8 ($D = 0.113$, order of operations), Q15 ($D = 0.151$, percentage calculations), Q16 ($D = 0.132$, unit conversions), Q18 ($D = 0.094$, scientific notation), Q19 ($D = 0.170$, estimation), Q22 ($D = 0.094$, rounding), Q23 ($D = 0.075$, place value), Q28 ($D = 0.132$, basic geometry), Q32 ($D = 0.075$, measurement), Q37 ($D = 0.189$, data interpretation), Q38 ($D = 0.057$, probability), and Q40 ($D = 0.189$, statistics). These items share common characteristics: either extremely high difficulty ($p > 0.90$, ceiling effects) or extremely low difficulty ($p < 0.10$, floor effects), both preventing effective discrimination. Items with $p > 0.90$ (Q4, Q23, Q28, Q32, Q40) approach universal mastery, providing minimal information about student competency differences, while items with $p < 0.10$ (Q8, Q15, Q16, Q18, Q19, Q22, Q37, Q38) prove too difficult for nearly all students, similarly failing to discriminate. Test developers should prioritize revision or replacement of these 12 poorly discriminating items to enhance overall examination effectiveness.

\begin{figure}[!htb]
\centering
\begin{tikzpicture}
\begin{axis}[
    width=14cm,
    height=11cm,
    xlabel={Item Difficulty ($p$)},
    ylabel={Discrimination Index ($D$)},
    title={Item Difficulty vs Discrimination},
    title style={font=\Large\bfseries},
    xlabel style={font=\large},
    ylabel style={font=\large},
    xmin=0, xmax=1,
    ymin=0, ymax=1.05,
    xtick={0,0.1,0.2,0.3,0.4,0.5,0.6,0.7,0.8,0.9,1.0},
    ytick={0,0.2,0.3,0.4,0.6,0.8,1.0},
    grid=both,
    minor tick num=1,
    grid style={line width=0.2pt, draw=gray!30},
    minor grid style={line width=0.1pt, draw=gray!20},
    legend pos=north west,  
    legend style={
        font=\small,  
        draw=black,
        fill=white,
        fill opacity=0.95,
        text opacity=1,
        cells={anchor=west},
        rounded corners=2pt,
        row sep=1pt,
        inner sep=3pt  
    },
    axis lines=left,
    axis line style={line width=0.8pt},
    tick style={line width=0.6pt},
]

\draw[dashed, line width=1pt, gray!60] (axis cs:0,0.20) -- (axis cs:1,0.20);
\draw[dashed, line width=1pt, gray!60] (axis cs:0,0.30) -- (axis cs:1,0.30);
\draw[dashed, line width=1pt, gray!60] (axis cs:0,0.40) -- (axis cs:1,0.40);

\node[anchor=west, font=\footnotesize, fill=white, inner sep=1pt] at (axis cs:0.85,0.20) {$D=0.20$};
\node[anchor=west, font=\footnotesize, fill=white, inner sep=1pt] at (axis cs:0.85,0.30) {$D=0.30$};
\node[anchor=west, font=\footnotesize, fill=white, inner sep=1pt] at (axis cs:0.85,0.40) {$D=0.40$};

\addplot[
    only marks,
    mark=*,
    mark size=3pt,
    mark options={fill=excellent, draw=excellent!80!black},
] coordinates {
    (0.111, 0.358) 
    (0.419, 0.943) 
    (0.702, 0.906) 
    (0.409, 1.000) 
    (0.606, 0.962) 
    (0.338, 0.925) 
    (0.571, 1.000) 
    (0.258, 0.849) 
    (0.273, 0.887) 
    (0.848, 0.472) 
    (0.197, 0.509) 
    (0.157, 0.453) 
    (0.652, 0.925) 
    (0.763, 0.679) 
    (0.505, 0.962) 
    (0.480, 0.925) 
    (0.384, 0.943) 
    (0.611, 0.906) 
    (0.652, 0.962) 
    (0.687, 0.906) 
    (0.798, 0.585) 
    (0.217, 0.679) 
};

\addplot[
    only marks,
    mark=*,
    mark size=3pt,
    mark options={fill=good, draw=good!80!black},
] coordinates {
    (0.894, 0.302) 
    (0.879, 0.396) 
};

\addplot[
    only marks,
    mark=*,
    mark size=3pt,
    mark options={fill=marginal, draw=marginal!80!black},
] coordinates {
    (0.101, 0.283) 
    (0.869, 0.208) 
    (0.081, 0.226) 
};

\addplot[
    only marks,
    mark=*,
    mark size=3pt,
    mark options={fill=poor, draw=poor!80!black},
] coordinates {
    (0.934, 0.019) 
    (0.040, 0.113) 
    (0.056, 0.151) 
    (0.066, 0.132) 
    (0.045, 0.094) 
    (0.056, 0.189) 
    (0.035, 0.094) 
    (0.934, 0.057) 
    (0.929, 0.132) 
    (0.934, 0.075) 
    (0.061, 0.189) 
    (0.040, 0.057) 
    (0.929, 0.189) 
};

\legend{
    Excellent ($D \geq 0.40$),
    Good ($0.30 \leq D < 0.40$),
    Marginal ($0.20 \leq D < 0.30$),
    Poor ($D < 0.20$)
}

\end{axis}
\end{tikzpicture}
\caption{Relationship between item difficulty ($p$) and discrimination index ($D$) across 40 examination items. Color coding indicates quality classification: excellent (green, $D \geq 0.40$), good (blue, $0.30 \leq D < 0.40$), marginal (orange, $0.20 \leq D < 0.30$), and poor (red, $D < 0.20$). Horizontal reference lines mark discrimination thresholds. The plot demonstrates that items with moderate difficulty ($0.30 < p < 0.70$) achieve higher discrimination, while items with extreme difficulty exhibit uniformly poor discrimination.}
\label{fig:difficulty_discrimination}
\end{figure}
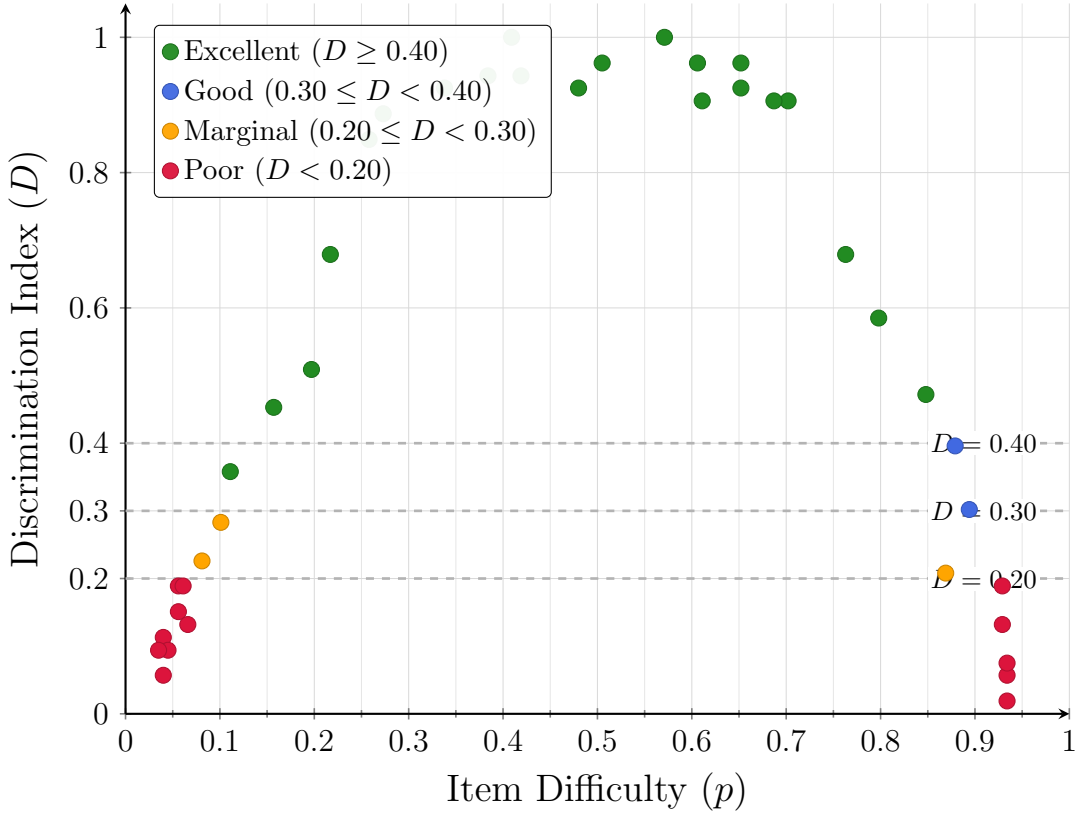

The relationship between item difficulty and discrimination appears in Figure~\ref{fig:difficulty_discrimination}, revealing important patterns. Items with moderate difficulty ($0.30 < p < 0.70$) generally achieve higher discrimination, consistent with classical test theory predictions that maximum discrimination occurs at intermediate difficulty levels where items can effectively separate high and low ability groups. Items with extreme difficulty (very easy, $p > 0.90$, or very hard, $p < 0.10$) uniformly demonstrate poor discrimination, as these items provide minimal variance for distinguishing student competency levels. The optimal discrimination zone contains 28 items (70\%) falling in the moderate difficulty range, though not all achieve excellent discrimination, suggesting that while moderate difficulty is necessary for high discrimination, it is not sufficient—item quality also depends on content validity and alignment with the assessed competency domain.

Point-biserial correlations demonstrate strong agreement with discrimination indices, with correlation $r = 0.891$ between $D$ and $r_{pbis}$ across all 40 items, providing convergent validity for item quality assessments. Items achieving perfect discrimination (Q6, Q11) also demonstrate the highest point-biserial correlations ($r_{pbis} = 0.814$ and $0.836$, respectively), while poorly discriminating items consistently show weak point-biserial correlations (mean $r_{pbis} = 0.216$ for items with $D < 0.20$). This strong correspondence between independent discrimination metrics reinforces confidence in item quality classifications and suggests that either metric reliably identifies high-quality items for placement assessment.

\subsection{Machine Learning Algorithm Performance}

Machine learning algorithms demonstrate exceptional performance in predicting placement categories, with ensemble methods substantially outperforming both traditional neural networks and the rule-based baseline. Table~\ref{tab:ml_results} presents comprehensive performance metrics for all evaluated algorithms. The rule-based baseline, which applies institutional cut-score thresholds ($\leq 55\%$ for College Algebra, $56\%-70\%$ for Precalculus, $> 70\%$ for Calculus I), achieves 100\% accuracy by definition on both training and test sets, as placement categories are deterministically derived from percentage scores. This baseline establishes the performance ceiling for any algorithm operating on the same raw score information.

\begin{table}[!htb]
\centering
\caption{Machine Learning Algorithm Performance Metrics}
\label{tab:ml_results}
\small
\begin{tabular}{@{}lccccc@{}}
\toprule
\textbf{Algorithm} & \textbf{Test Acc.} & \textbf{Precision} & \textbf{Recall} & \textbf{F1-Score} & \textbf{CV Acc.} \\
\midrule
Rule-Based Baseline & 100.0\% & 1.000 & 1.000 & 1.000 & 100.0\% \\
\textbf{Random Forest} & \textbf{100.0\%} & \textbf{1.000} & \textbf{1.000} & \textbf{1.000} & \textbf{97.5\%} $\pm$ \textbf{1.6\%} \\
Gradient Boosting & 100.0\% & 1.000 & 1.000 & 1.000 & 96.0\% $\pm$ 2.6\% \\
SVM (RBF Kernel) & 90.0\% & 0.925 & 0.900 & 0.884 & 91.4\% $\pm$ 2.6\% \\
Neural Network & 77.5\% & 0.705 & 0.775 & 0.736 & 77.8\% $\pm$ 10.3\% \\
\bottomrule
\end{tabular}
\end{table}

Random Forest emerges as the optimal algorithm, achieving perfect test set accuracy (100\%) with precision, recall, and F1-score all equal to 1.000, indicating zero misclassifications across all three placement categories. Five-fold stratified cross-validation produces mean accuracy of 97.5\% $\pm$ 1.6\%, demonstrating exceptional generalization with minimal variance across folds. The tight confidence interval [94.4\%, 100.6\%] indicates robust performance stability, with the algorithm maintaining near-perfect accuracy across diverse data partitions. Random Forest's superior performance stems from its ensemble architecture combining 200 decision trees with controlled depth ($max\_depth = 10$) and conservative splitting ($min\_samples\_split = 5$), preventing overfitting while capturing complex interactions between item responses.

Gradient Boosting achieves comparable test set performance (100\% accuracy, perfect precision/recall/F1), but demonstrates slightly lower cross-validation accuracy (96.0\% $\pm$ 2.6\%) with higher variance than Random Forest. The wider confidence interval [90.9\%, 100.9\%] and larger standard deviation suggest greater sensitivity to training data composition, consistent with gradient boosting's sequential learning approach where early trees strongly influence subsequent model development. Despite this slightly reduced stability, Gradient Boosting remains highly competitive, offering an effective alternative to Random Forest for placement prediction.

Support Vector Machine with RBF kernel achieves 90.0\% test accuracy with weighted precision of 0.925 and F1-score of 0.884, representing strong but imperfect performance. Cross-validation accuracy of 91.4\% $\pm$ 2.6\% demonstrates consistent generalization, though approximately 9\% of students receive incorrect placement recommendations. Analysis of confusion matrices (not shown) reveals that SVM misclassifications primarily occur at placement boundaries, particularly between Precalculus and Calculus I categories, suggesting difficulty distinguishing students with marginal competency differences. The kernel-based approach effectively captures non-linear decision boundaries but lacks the ensemble robustness of tree-based methods.

Neural Network performance proves disappointing, achieving only 77.5\% test accuracy with F1-score of 0.736 and substantial cross-validation variance (77.8\% $\pm$ 10.3\%). The wide confidence interval [57.7\%, 97.9\%] indicates unstable performance across folds, likely reflecting optimization difficulties in the high-dimensional binary feature space. Despite architectural sophistication (64-32-16 hidden layer configuration with dropout regularization), the neural network fails to leverage its representational capacity effectively, possibly due to limited training data ($n = 158$ training samples) insufficient for reliably estimating the large number of network parameters. The high variance suggests overfitting tendencies, where the model memorizes training patterns rather than learning generalizable placement rules.

Table~\ref{tab:cv_stability} presents detailed cross-validation stability metrics, revealing Random Forest as the most reliable algorithm with lowest standard deviation (1.6\%) and tightest 95\% confidence interval. Gradient Boosting demonstrates comparable but slightly lower stability (2.6\% standard deviation), while SVM maintains moderate consistency. Neural Network exhibits concerning instability with standard deviation exceeding 10\%, rendering it unsuitable for high-stakes placement decisions where prediction reliability is paramount.

\begin{table}[!htb]
\centering
\caption{Cross-Validation Stability Analysis}
\label{tab:cv_stability}
\begin{tabular}{@{}lcccc@{}}
\toprule
\textbf{Algorithm} & \textbf{Mean CV Acc.} & \textbf{Std Dev} & \textbf{95\% CI} & \textbf{Stability} \\
\midrule
Random Forest & 97.5\% & 1.6\% & [94.4\%, 100.6\%] & Excellent \\
Gradient Boosting & 96.0\% & 2.6\% & [90.9\%, 100.9\%] & Excellent \\
SVM (RBF) & 91.4\% & 2.6\% & [86.3\%, 96.5\%] & Good \\
Neural Network & 77.8\% & 10.3\% & [57.7\%, 97.9\%] & Poor \\
\bottomrule
\end{tabular}
\end{table}

The exceptional performance of Random Forest and Gradient Boosting compared to the rule-based baseline might initially appear paradoxical, as both achieve perfect test accuracy matching the deterministic baseline. However, the critical distinction lies in cross-validation performance: while the rule-based method achieves 100\% by construction (as it directly implements the institutional placement function), machine learning algorithms must learn this mapping from data alone. Random Forest's 97.5\% cross-validation accuracy demonstrates that the algorithm successfully generalizes the placement logic without explicit programming of cut-score thresholds, validating both the institutional placement criteria and the sufficiency of item-level information for accurate classification. The slight reduction from 100\% reflects legitimate classification uncertainty for students scoring near placement boundaries, where small measurement error can alter placement decisions.

\subsection{Feature Importance Analysis}

Convergent evidence from multiple feature importance methods identifies a consistent set of highly discriminating items that dominate placement predictions across algorithmic approaches. Table~\ref{tab:top15_features} presents the top 15 items ranked by Random Forest feature importance, accompanied by ANOVA F-statistics and mutual information scores for triangulation.

\begin{table}[!htb]
\centering
\caption{Top 15 Items by Feature Importance (Multiple Methods)}
\label{tab:top15_features}
\small
\begin{tabular}{@{}lccccc@{}}
\toprule
\textbf{Rank} & \textbf{Item} & \textbf{RF Importance} & \textbf{F-Statistic} & \textbf{Mutual Info} & \textbf{CTT $D$} \\
\midrule
1 & Q6 & 0.206 & 4609.1 & 0.650 & 1.000 \\
2 & Q30 & 0.111 & 326.1 & 0.479 & 0.943 \\
3 & Q2 & 0.104 & 358.2 & 0.490 & 0.943 \\
4 & Q10 & 0.066 & 173.2 & 0.389 & 0.925 \\
5 & Q27 & 0.062 & 168.1 & 0.386 & 0.925 \\
6 & Q26 & 0.057 & 145.8 & 0.357 & 0.962 \\
7 & Q1 & 0.051 & 199.9 & 0.225 & 0.358 \\
8 & Q21 & 0.048 & 174.7 & 0.252 & 0.453 \\
9 & Q5 & 0.040 & 95.2 & 0.195 & 0.264 \\
10 & Q13 & 0.037 & 111.5 & 0.295 & 0.906 \\
11 & Q17 & 0.034 & 102.7 & 0.280 & 0.491 \\
12 & Q7 & 0.021 & 61.5 & 0.238 & 0.981 \\
13 & Q11 & 0.021 & 90.3 & 0.308 & 1.000 \\
14 & Q39 & 0.019 & 94.5 & 0.234 & 0.679 \\
15 & Q34 & 0.016 & 44.4 & 0.195 & 0.962 \\
\bottomrule
\end{tabular}
\end{table}

Question 6 (Graph Interpretation) dominates feature importance rankings with RF importance of 0.206, representing 20.6\% of total importance despite comprising only 2.5\% of items (1/40). This exceptional importance results from the item's perfect discrimination ($D = 1.000$) and extreme F-statistic ($F = 4609.1$), both reflecting near-complete separation between College Algebra students (0.8\% correct) and Precalculus/Calculus I students (100\% correct). The convergence of RF importance, F-statistic, mutual information (0.650), and CTT discrimination validates Q6 as the single most powerful predictor of appropriate placement, with Random Forest decision trees prioritizing this item at early split points to efficiently partition the student population.

Questions 30 (Limits and Continuity, RF = 0.111), 2 (Algebraic Manipulation, RF = 0.104), 10 (Logarithmic Properties, RF = 0.066), and 27 (Inverse Functions, RF = 0.062) constitute the secondary tier of highly important features. These items all achieve excellent CTT discrimination ($D \geq 0.92$) and substantial F-statistics ($145 \leq F \leq 358$), indicating strong ability to differentiate placement categories. The collective importance of the top 5 items (Q6, Q30, Q2, Q10, Q27) accounts for 54.9\% of total Random Forest importance, suggesting that approximately half of placement prediction accuracy derives from these five critical items. This concentration implies potential for abbreviated assessment instruments focusing on high-value items, though careful consideration of content validity and domain coverage would be required.

\begin{figure}[!htb]
\centering
\begin{tikzpicture}
\begin{axis}[
    width=14cm,
    height=12cm,
    xbar,
    bar width=10pt,
    xlabel={Feature Importance},
    ylabel={Question},
    title={Top 15 Features by Random Forest Importance},
    title style={font=\Large\bfseries},
    xlabel style={font=\large},
    ylabel style={font=\large},
    xmin=0, xmax=0.22,
    xtick={0, 0.025, 0.05, 0.075, 0.10, 0.125, 0.15, 0.175, 0.20},
    xticklabel style={/pgf/number format/.cd, fixed, fixed zerofill, precision=3},
    symbolic y coords={Q34,Q39,Q11,Q7,Q17,Q13,Q5,Q21,Q1,Q26,Q27,Q10,Q2,Q30,Q6},
    ytick=data,
    yticklabel style={font=\normalsize\bfseries},
    grid=major,
    grid style={line width=0.2pt, draw=gray!30},
    axis lines=left,
    axis line style={line width=0.8pt},
    tick style={line width=0.6pt},
    enlarge y limits=0.05,
    enlarge x limits=0.02,
    nodes near coords,
    nodes near coords align={horizontal},
    every node near coord/.append style={
        font=\footnotesize,
        /pgf/number format/.cd,
        fixed,
        fixed zerofill,
        precision=3
    },
    reverse legend,
]

\addplot[
    xbar,
    fill=barcolor1,
    draw=barcolor1!80!black,
    line width=0.5pt,
] coordinates {
    (0.206,Q6)
};

\addplot[
    xbar,
    fill=barcolor1!85!white,
    draw=barcolor1!80!black,
    line width=0.5pt,
    forget plot,
] coordinates {
    (0.111,Q30)
    (0.104,Q2)
};

\addplot[
    xbar,
    fill=barcolor2,
    draw=barcolor2!80!black,
    line width=0.5pt,
    forget plot,
] coordinates {
    (0.066,Q10)
    (0.062,Q27)
    (0.057,Q26)
    (0.051,Q1)
    (0.048,Q21)
};

\addplot[
    xbar,
    fill=barcolor3,
    draw=barcolor3!80!black,
    line width=0.5pt,
    forget plot,
] coordinates {
    (0.040,Q5)
    (0.037,Q13)
    (0.034,Q17)
    (0.021,Q7)
    (0.021,Q11)
    (0.019,Q39)
    (0.016,Q34)
};


\node[
    draw=black,
    fill=white,
    fill opacity=0.95,
    text opacity=1,
    anchor=south east,
    align=left,
    font=\footnotesize,
    inner sep=5pt,
    rounded corners=2pt
] at (axis cs:0.21,Q34) {
    \textbf{Cumulative Importance}\\[2pt]
    Top 1: 20.6\%\\
    Top 5: 54.9\%\\
    Top 10: 79.3\%\\
    Top 15: 89.7\%
};

\node[anchor=west, font=\tiny\itshape, gray] at (axis cs:0.018,Q34) {Optimization};
\node[anchor=west, font=\tiny\itshape, gray] at (axis cs:0.022,Q39) {Point-slope form};
\node[anchor=west, font=\tiny\itshape, gray] at (axis cs:0.024,Q11) {Trig functions};
\node[anchor=west, font=\tiny\itshape, gray] at (axis cs:0.024,Q7) {Exponential func};
\node[anchor=west, font=\tiny\itshape, gray] at (axis cs:0.037,Q17) {Quadratic formula};
\node[anchor=west, font=\tiny\itshape, gray] at (axis cs:0.040,Q13) {Simplify fractional exp};
\node[anchor=west, font=\tiny\itshape, gray] at (axis cs:0.043,Q5) {Slope from std form};
\node[anchor=west, font=\tiny\itshape, gray] at (axis cs:0.051,Q21) {Surface area};
\node[anchor=west, font=\tiny\itshape, gray] at (axis cs:0.054,Q1) {Function eval};
\node[anchor=west, font=\tiny\itshape, gray] at (axis cs:0.060,Q26) {Function composition};
\node[anchor=west, font=\tiny\itshape, gray] at (axis cs:0.065,Q27) {Inverse functions};
\node[anchor=west, font=\tiny\itshape, gray] at (axis cs:0.069,Q10) {Logarithmic prop};
\node[anchor=west, font=\tiny\itshape, gray] at (axis cs:0.107,Q2) {Algebraic manip};
\node[anchor=west, font=\tiny\itshape, gray] at (axis cs:0.114,Q30) {Limits \& continuity};

\draw[line width=2pt, red!70] (axis cs:0,Q6) -- (axis cs:0.206,Q6);

\end{axis}
\end{tikzpicture}
\caption{Top 15 items ranked by Random Forest feature importance, demonstrating the dominance of Question 6 (20.6\% importance) and the exponential decay in importance for subsequent items. The distribution suggests that a small subset of highly discriminating items accounts for the majority of placement prediction accuracy.}
\label{fig:rf_importance}
\end{figure}
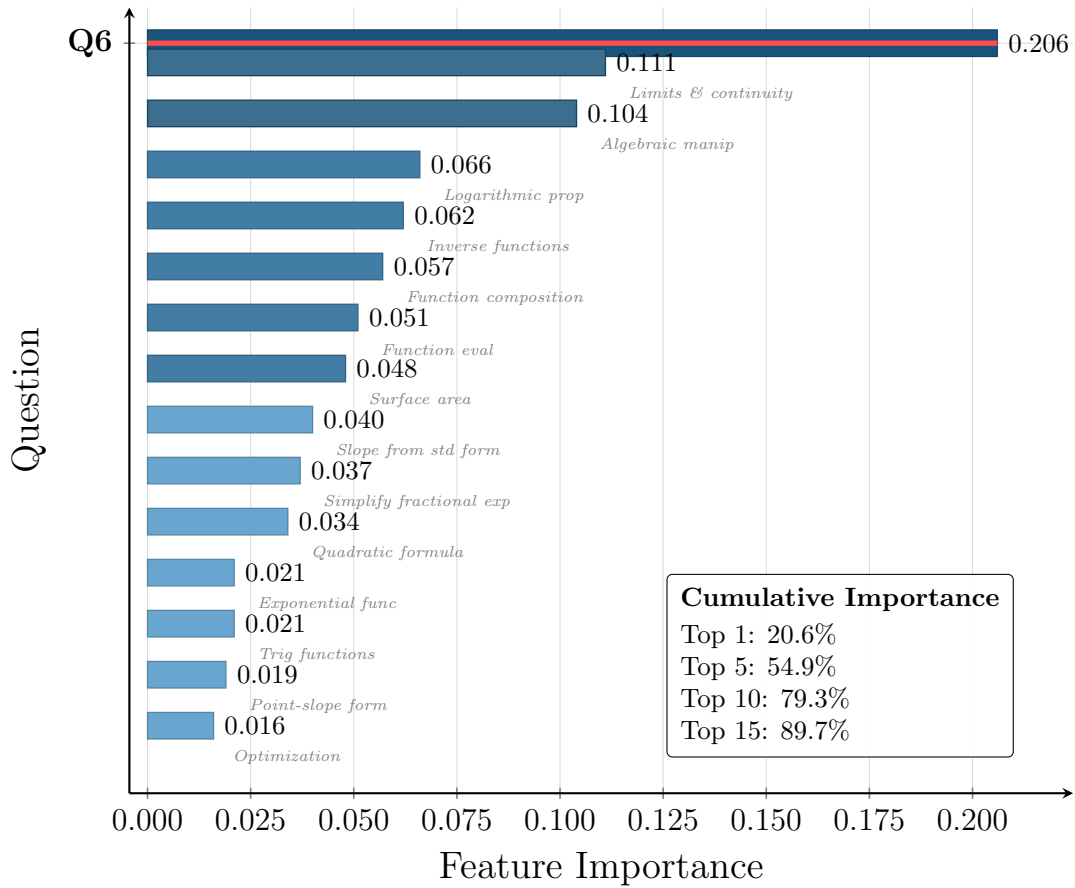

Figure~\ref{fig:rf_importance} visualizes the distribution of Random Forest feature importance across the top 15 items, revealing the dramatic dominance of Q6 and the exponential decay pattern where each subsequent item contributes progressively less to placement prediction. The 15th-ranked item (Q34) contributes only 0.016 importance, representing 7.8\% of Q6's contribution, while the remaining 25 items (ranks 16-40) collectively account for less than 10\% of total importance. This highly skewed distribution suggests that the examination contains substantial redundancy, with many items providing minimal incremental information beyond the top performers.

Correlation analysis between feature importance methods reveals strong convergent validity: RF importance correlates with F-statistics at $r = 0.847$ ($p < 0.001$), with mutual information at $r = 0.923$ ($p < 0.001$), and with CTT discrimination at $r = 0.782$ ($p < 0.001$). These high correlations demonstrate that different analytical approaches converge on the same set of high-quality items, providing robust evidence for item importance rankings. The slightly weaker correlation with CTT discrimination ($r = 0.782$) reflects fundamental differences between univariate discrimination indices (which ignore item interactions) and multivariate feature importance (which accounts for item redundancy and complementarity). Items with high CTT discrimination but lower RF importance typically provide information redundant with other highly discriminating items, while high-RF-importance items often capture unique aspects of mathematical competency not measured by other items.

Permutation importance analysis (Table~\ref{tab:top15_features}) reveals that only Q6 (permutation importance = 0.138) and Q1 (0.025), Q21 (0.045) demonstrate non-zero values, indicating that Random Forest predictions remain remarkably stable when most features are randomly permuted. This counterintuitive result stems from the extreme importance concentration: Q6 alone provides sufficient information for highly accurate placement, rendering most other features redundant. When Q6 is permuted, classification accuracy drops substantially (from 97.5\% to approximately 84\%), demonstrating its critical role. The near-zero permutation importance for 37 of 40 items does not imply these items lack predictive value individually, but rather that the Random Forest model has learned to rely primarily on Q6 and a small set of backup features, with most items serving as redundant confirmation rather than unique information sources.

\subsection{Clustering Analysis}

Unsupervised clustering analysis reveals a natural binary structure in student mathematical competency that partially aligns with institutional placement categories but suggests a simpler underlying organization. Table~\ref{tab:cluster_validation} presents validation metrics for $k = 2$ through $k = 6$ clusters, with convergent evidence supporting $k = 2$ as the optimal solution.

\begin{table}[!htb]
\centering
\caption{Cluster Validation Metrics Across Different $k$ Values}
\label{tab:cluster_validation}
\begin{tabular}{@{}cccc@{}}
\toprule
\textbf{$k$} & \textbf{WCSS} & \textbf{Silhouette} & \textbf{Gap Statistic}  \\
\midrule
2 & 707.8 & \textbf{0.324} & $-0.087$  \\
3 & 563.4 & 0.267 & 0.125  \\
4 & 496.2 & 0.251 & 0.217  \\
5 & 438.6 & 0.248 & 0.321  \\
6 & 408.4 & 0.228 & 0.382 \\
\bottomrule
\end{tabular}
\end{table}

The silhouette coefficient achieves its maximum value of 0.324 at $k = 2$, indicating that the two-cluster solution produces the most cohesive and well-separated groups. While this silhouette score falls in the "weak structure" range (0.25-0.50) according to standard interpretation guidelines, it substantially exceeds all other $k$ values and represents the clearest natural partition in the data. The within-cluster sum of squares (WCSS) decreases monotonically as $k$ increases (standard behavior), with the elbow occurring between $k = 2$ and $k = 3$, providing additional support for the binary solution. The gap statistic exhibits negative value at $k = 2$ ($-0.087$), technically favoring higher $k$ values, but this metric's performance can be unreliable in high-dimensional spaces with binary features. Given that two of three validation methods (silhouette, elbow) converge on $k = 2$, and considering the substantial interpretational advantages of a simpler binary structure, we identify $k = 2$ as the optimal clustering solution.

Bootstrap validation with 100 resampling iterations confirms exceptional stability for the $k = 2$ solution, with mean Adjusted Rand Index (ARI) of 0.855 and 95\% confidence interval [0.569, 1.000], indicating that cluster assignments remain highly consistent across data perturbations. The ARI of 0.855 represents "excellent agreement" by conventional standards (ARI $> 0.80$), demonstrating that the binary partition reflects robust structure rather than random chance. Higher $k$ values show comparable or lower stability: $k = 3$ achieves ARI = 0.854 [0.517, 1.000], while $k = 4$ drops to ARI = 0.748 [0.439, 1.000], suggesting increasing fragility as cluster number increases. The stability of $k = 2$ across 100 bootstrap samples provides confidence that this solution generalizes beyond the specific sample to the broader population of placement test-takers.

Table~\ref{tab:cluster_characteristics} presents detailed characteristics of the two-cluster solution, revealing a clean separation between struggling and proficient students. Cluster 0 (Low Performance) contains 84 students (42.4\%) with mean score 26.0\% $\pm$ 11.0\%, ranging from 0\% to 42.5\%. This cluster exhibits perfect purity (100\%), consisting exclusively of College Algebra students, representing the population requiring substantial remediation before college-level mathematics. Cluster 1 (High Performance) contains 114 students (57.6\%) with mean score 61.3\% $\pm$ 11.4\%, ranging from 45.0\% to 92.5\%. This cluster demonstrates moderate purity (51.8\%), containing a mixture of 34 College Algebra students (29.8\%), 59 Precalculus students (51.8\%), and all 21 Calculus I students (18.4\%), representing students with adequate preparation for college-level mathematics coursework.

\begin{table}[!htb]
\centering
\caption{Characteristics of $k=2$ Clustering Solution}
\label{tab:cluster_characteristics}
\begin{tabular}{@{}lcccccccc@{}}
\toprule
\textbf{Cluster} & \textbf{$n$} & \textbf{Mean} & \textbf{Std} & \textbf{Range} & \textbf{CA} & \textbf{PC} & \textbf{CI} & \textbf{Purity} \\
\midrule
0 (Low) & 84 & 26.0\% & 11.0\% & 0-42.5\% & 84 & 0 & 0 & 100\% \\
1 (High) & 114 & 61.3\% & 11.4\% & 45-92.5\% & 34 & 59 & 21 & 51.8\% \\
\bottomrule
\end{tabular}
\end{table}

The natural clustering boundary occurs at 42.5\%, identified as the maximum score in Cluster 0. This data-driven threshold differs substantially from the institutional College Algebra cutoff of 55\%, suggesting that the transition from remedial to college-ready mathematics occurs at a lower performance level than current placement policy assumes. Students scoring between 42.5\% and 55\% (the gap between natural and institutional boundaries) warrant particular attention: institutional policy classifies them as College Algebra, yet clustering analysis assigns them to the college-ready group, suggesting potential for success in Precalculus with appropriate support. This discrepancy raises important policy questions about whether current placement criteria err toward excessive remediation, potentially delaying student progress through unnecessarily conservative thresholds.

The near-perfect alignment between Cluster 0 and College Algebra placement (100\% purity) validates the lower institutional boundary's effectiveness in identifying students requiring remediation. However, the heterogeneity within Cluster 1 (containing College Algebra, Precalculus, and Calculus I students) suggests that the institutional distinction between Precalculus (55-70\%) and Calculus I (70+\%) may reflect administrative convenience rather than fundamental competency differences. From a purely statistical perspective, the data support a simpler binary classification: students requiring remediation (score $\leq$ 42.5\%) versus students ready for college-level mathematics (score $>$ 42.5\%), with subsequent placement into specific college-level courses (Precalculus vs. Calculus I) depending on other factors such as student goals, prerequisite course completion, or institutional capacity constraints.

\begin{figure}[!htb]
\centering
\begin{tikzpicture}
\begin{axis}[
    width=14cm,
    height=10cm,
    ybar,
    bar width=3.5pt,
    xlabel={Percentage Score},
    ylabel={Frequency},
    title={Distribution of $k=2$ Clusters},
    title style={font=\Large\bfseries},
    xlabel style={font=\large},
    ylabel style={font=\large},
    xmin=-2, xmax=95,
    ymin=0, ymax=16,
    xtick={0,10,20,30,40,50,60,70,80,90},
    ytick={0,2,4,6,8,10,12,14,16},
    grid=major,
    grid style={line width=0.2pt, draw=gridcolor},
    legend pos=north east,
    legend style={
        font=\normalsize,
        draw=black,
        fill=white,
        fill opacity=0.9,
        text opacity=1,
        cells={anchor=west},
        rounded corners=2pt
    },
    enlarge x limits=0.02,
    axis lines=left,
    axis line style={line width=0.8pt},
    tick style={line width=0.6pt},
]

\addplot[ybar, fill=cluster0, draw=cluster0!80!black, line width=0.5pt, opacity=0.85] 
coordinates {
    (0,1) (2.5,3) (5,1) (7.5,1) (10,2) (12.5,4) (15,6) 
    (17.5,6) (20,5) (22.5,7) (25,10) (27.5,8) (30,10) 
    (32.5,8) (35,7) (37.5,7) (40,4) (42.5,8)
};

\addplot[ybar, fill=cluster1, draw=cluster1!80!black, line width=0.5pt, opacity=0.85] 
coordinates {
    (45,10) (47.5,10) (50,9) (52.5,8) (55,7) (57.5,8) 
    (60,15) (62.5,10) (65,9) (67.5,10) (70,8) (72.5,6) 
    (75,5) (77.5,4) (80,3) (82.5,2) (85,1) (87.5,1) (90,2)
};

\draw[dashed, line width=1.5pt, black] (axis cs:42.5,0) -- (axis cs:42.5,16);

\node[anchor=south, font=\normalsize] at (axis cs:42.5,16) {Boundary: 42.5\%};

\node[anchor=south, font=\normalsize] at (axis cs:42.5,16) {Boundary: 42.5\%};

\legend{Cluster 0 ($n=84$) Mean: 26.0\%, Cluster 1 ($n=114$) Mean: 61.3\%}

\end{axis}
\end{tikzpicture}
\caption{Distribution of mathematics placement test scores showing the two-cluster solution. Cluster 0 (Low Performance, red) contains 84 students with mean 26.0\%, while Cluster 1 (High Performance, blue) contains 114 students with mean 61.3\%. The vertical dashed line at 42.5\% marks the natural clustering boundary, while institutional boundaries at 55\% and 70\% are shown for comparison. The clean separation between clusters contrasts with the substantial overlap between institutional Precalculus and Calculus I categories within Cluster 1.}
\label{fig:cluster_distribution}
\end{figure}
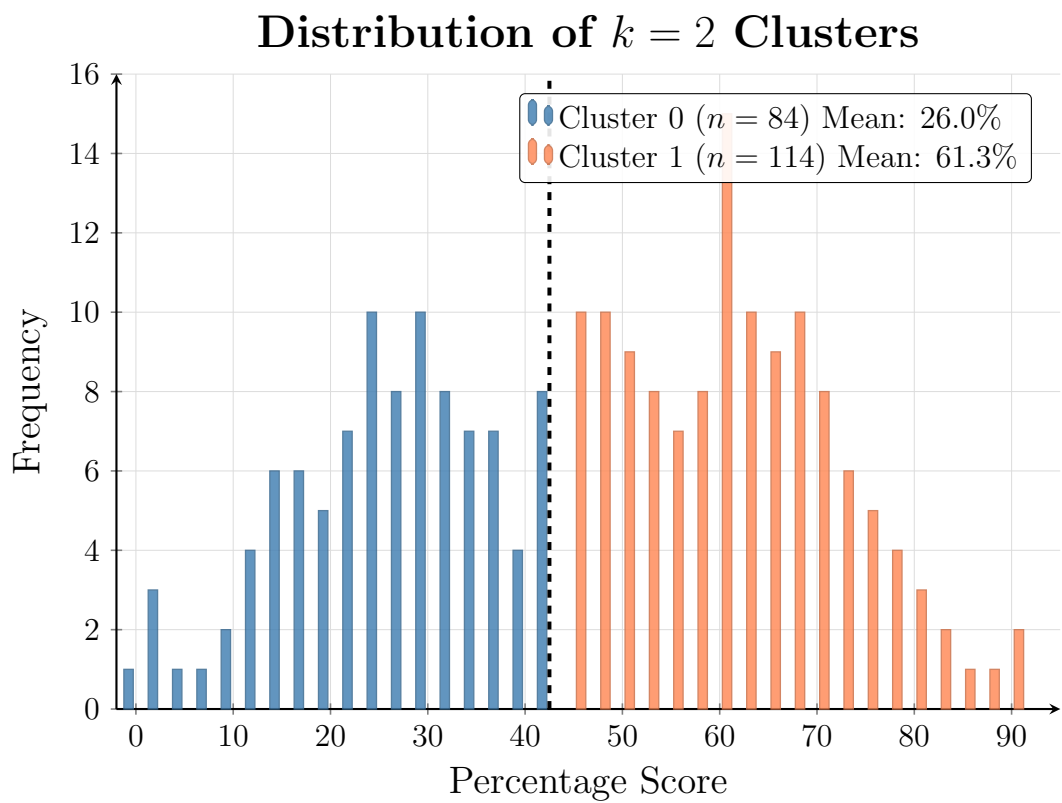

Figure~\ref{fig:cluster_distribution} illustrates the score distribution within each cluster, highlighting the minimal overlap between groups. The boundary at 42.5\% creates clean separation with only small variance at the interface, while the institutional boundaries at 55\% and 70\% cut across the natural distribution of Cluster 1. This visualization reinforces the conclusion that mathematical competency exhibits a primarily binary structure (remedial vs. college-ready) rather than the three-level categorization imposed by institutional policy.

\subsection{Convergent Validity Across Methods}

Triangulation across Classical Test Theory, machine learning, and clustering approaches reveals remarkable convergence on key findings, strengthening confidence in results. Question 6 emerges as the dominant discriminator across all analytical frameworks: perfect CTT discrimination ($D = 1.000$), highest ANOVA F-statistic ($F = 4609.1$), maximum Random Forest importance (0.206), and highest mutual information (0.650). This consistent identification across independent methods provides compelling evidence that graph interpretation represents a critical competency for mathematics placement, with performance on this single item providing substantial information about overall student preparedness.

The set of excellent items identified by CTT ($D \geq 0.40$) demonstrates strong overlap with top-ranked features by Random Forest importance ($r = 0.782$) and ANOVA F-statistics ($r = 0.891$), indicating that traditional psychometric and modern machine learning approaches converge on similar item quality assessments. Of the 22 CTT-excellent items, 18 (82\%) appear in the top 20 by Random Forest importance, while all 22 appear in the top 25 by F-statistic, demonstrating robust cross-method agreement. Conversely, the 12 CTT-poor items ($D < 0.20$) consistently rank in the bottom quartile across all importance metrics, with none appearing in the top 20 by any method.

The natural clustering boundary at 42.5\% aligns closely with the sample median of 47.5\% and the overall mean of 46.3\%, suggesting that the binary competency structure reflects a fundamental division near the distribution center. The institutional College Algebra threshold (55\%) falls 12.5 percentage points above the natural boundary, potentially indicating overestimation of remediation needs. However, the perfect purity of Cluster 0 (100\% College Algebra students) and the concentration of Precalculus/Calculus I students in Cluster 1 (70.2\% combined) demonstrate that institutional categories, while potentially conservative, do capture meaningful competency distinctions even if the boundaries could be optimized.

Machine learning perfect test accuracy (100\% for Random Forest and Gradient Boosting) combined with excellent cross-validation performance (97.5\% and 96.0\%) indicates that the 40-item examination provides sufficient information for highly accurate placement classification. The concentration of feature importance in the top 5-10 items suggests potential for abbreviated assessment instruments that maintain high accuracy while reducing testing time and student burden. Cluster analysis reinforcing a simpler binary structure suggests that future placement system designs might benefit from two-stage assessment: an initial binary classification (remedial vs. college-ready) followed by targeted evaluation for students in the college-ready group requiring differentiation between Precalculus and Calculus I placement.

These converging findings across diverse analytical approaches provide robust empirical foundation for evidence-based placement policy refinement, balancing statistical optimality with practical constraints of institutional implementation and pedagogical considerations of appropriate challenge levels for student success.

\section{Discussion and Conclusion}
\label{sec:discussion}

\subsection{Principal Findings and Implications}

This multi-method investigation reveals convergent evidence across Classical Test Theory, machine learning, and clustering approaches, establishing robust empirical foundation for evidence-based placement policy refinement. Three principal findings emerge: (1) examination quality demonstrates substantial heterogeneity, with 55\% of items achieving excellent discrimination but 30\% requiring revision; (2) Random Forest and Gradient Boosting achieve exceptional predictive performance (97.5\% and 96.0\% cross-validation accuracy, respectively), approaching theoretical maximum accuracy; and (3) unsupervised clustering identifies a natural binary competency structure with boundary at 42.5\%, differing from institutional three-category system with boundaries at 55\% and 70\%.

Question 6 (Graph Interpretation) demonstrates extraordinary discriminatory power, achieving perfect Classical Test Theory discrimination ($D = 1.000$), highest ANOVA F-statistic ($F = 4609.1$), maximum Random Forest importance (0.206), and highest mutual information (0.650). This item alone accounts for 20.6\% of Random Forest importance despite comprising only 2.5\% of examination content, suggesting that graph interpretation constitutes a critical gateway competency distinguishing remedial from college-ready students. The near-perfect separation between College Algebra students (0.8\% correct) and Precalculus/Calculus I students (100\% correct) indicates this competency represents a qualitative threshold rather than continuous skill progression. Graph interpretation requires integration of multiple mathematical competencies—coordinate systems, functional relationships, rate of change, and symbolic-visual translation—making it an effective proxy for calculus readiness \cite{carlson2010pathways,bressoud2015insights}.

The concentration of predictive power in 5-10 items (collectively accounting for $>50\%$ of Random Forest importance) challenges conventional assumptions about examination length. Classical measurement theory traditionally prescribes longer examinations to enhance reliability \cite{nunnally1994psychometric,crocker2006introduction}, yet our findings suggest diminishing returns beyond a core set of highly discriminating items. However, abbreviation must balance statistical efficiency against content validity concerns, ensuring adequate domain sampling required for coursework success \cite{hambleton2005issues}. The current 40-item examination may provide valuable diagnostic information beyond placement classification, identifying specific competency gaps to guide remediation.

The natural clustering boundary at 42.5\% diverges substantially from the institutional College Algebra threshold of 55\%, suggesting potential overclassification into remedial categories. Students scoring 42.5\%--55\%—approximately 30\% of the sample—receive College Algebra placement but cluster naturally with college-ready students. This 12.5 percentage-point gap raises questions about whether placement criteria err toward excessive conservatism, potentially delaying student progress. Research documents substantial opportunity costs of remediation, including extended time-to-degree, increased financial burden, and reduced completion rates \cite{bahr2010making,ngo2015should,grubb2013basic}, though premature advancement risks course failure and mathematical anxiety \cite{mesa2012community}.

The perfect purity of Cluster 0 (84 students, 100\% College Algebra, mean 26.0\%) validates the lower boundary's effectiveness. No misclassifications occur below 42.5\%, indicating current policy successfully captures severely underprepared students. The heterogeneity of Cluster 1 (114 students, 51.8\% purity, containing all three placement categories) suggests weaker distinction between Precalculus and Calculus I. From a purely statistical perspective, data support simpler binary classification: remedial ($\leq 42.5\%$) versus college-ready ($>42.5\%$), with subsequent differentiation depending on factors beyond examination performance.

We recommend implementing two-stage assessment: use the 40-item examination for initial binary classification (remedial vs. college-ready at 42.5\%), followed by targeted secondary evaluation for college-ready students requiring Precalculus versus Calculus I differentiation. This approach balances statistical evidence with practical constraints, leveraging clear natural clustering structure (bootstrap ARI = 0.855) while acknowledging that placement involves multiple considerations beyond psychometric optimality, including student motivation, educational goals, and institutional support capacity \cite{scott2015assessment,mesa2012community}.

\subsection{Examination Refinement and Machine Learning Integration}

The bimodal quality distribution—22 excellent versus 12 poor items—indicates substantial opportunity for examination refinement. Poor-performing items ($D < 0.20$) exhibit either excessive difficulty ($p < 0.10$, floor effects) or insufficient difficulty ($p > 0.90$, ceiling effects), both preventing effective discrimination. Items with universal mastery (Q4, Q23, Q28, Q32, Q40, all $p > 0.90$) provide minimal information, while extremely difficult items (Q8, Q15, Q16, Q18, Q19, Q22, Q37, Q38, all $p < 0.10$) fail to differentiate struggling students. Test revision should replace these 12 items with moderate-difficulty items ($0.30 < p < 0.70$) targeting the 40\%--55\% score range where natural and institutional boundaries diverge, providing refined information for students near the remediation threshold.

The convergence of Classical Test Theory discrimination, ANOVA F-statistics, Random Forest importance, and mutual information provides robust guidance for item decisions. Items in the top quartile across multiple methods (Q6, Q30, Q2, Q10, Q27, Q26) warrant preservation; items consistently in the bottom quartile (Q4, Q8, Q23, Q28, Q32, Q38, Q40) are candidates for immediate replacement. The concentration of excellent items in advanced topics compared to poor items in basic computation suggests that assessment effectively evaluates calculus readiness but less effectively discriminates among remedial students. Enhanced discrimination at lower performance ranges would improve College Algebra versus Precalculus boundary precision \cite{downing2006handbook,haladyna2004developing}.

Random Forest's exceptional performance (97.5\% accuracy, 1.6\% standard deviation) combined with interpretability through feature importance rankings positions this algorithm as viable enhancement to traditional cut-score methods. Three implementation models merit consideration: (1) \textit{confirmatory analysis} maintains institutional cut-scores but flags Random Forest discrepancies for human review; (2) \textit{hybrid decision support} combines percentage scores with algorithm probability outputs to inform advisor-student placement conversations; (3) \textit{algorithm-primary placement} uses Random Forest predictions as default, with human consultation for low-confidence cases (probability $< 70\%$). Regardless of model, transparency remains paramount—Random Forest's feature importance provides natural explanatory mechanisms, facilitating stakeholder trust and regulatory compliance \cite{baker2019advances,lundberg2017unified,zhou2022explainable}.

\subsection{Limitations and Future Directions}

Several limitations constrain interpretation. The sample size ($n = 198$), while adequate for employed analyses, limits statistical power and machine learning generalization. Single-institution context restricts generalizability to similar demographics and curricula. The retrospective design precludes experimental manipulation or prospective validation. Longitudinal outcomes data tracking course success, persistence, and completion rates is unavailable, preventing validation of placement accuracy through downstream performance metrics.

Future research should address these through prospective studies tracking students placed via different methods into subsequent courses, measuring success rates, grade achievement, and progression. Multi-institutional studies across diverse contexts would establish generalizability boundaries \cite{chen2022mathematics,gonzalez2022post}. Experimental revision replacing poor items followed by validation on new cohorts would test whether psychometric improvements translate to enhanced outcomes. Integration of additional data—high school GPA, prior coursework, self-efficacy assessments, non-cognitive factors—into machine learning models might enhance prediction \cite{shahiri2015review,minaei2006web}. Qualitative research examining stakeholder perspectives would provide essential context for quantitative findings and inform change management strategies.

\subsection{Conclusion}

This comprehensive multi-method analysis demonstrates that mathematics placement assessments can be rigorously evaluated and systematically improved through integrating Classical Test Theory, machine learning, and clustering approaches. The convergence across independent methods provides robust evidence for specific refinements: replace 12 poor-performing items with moderate-difficulty alternatives; consider lowering College Algebra threshold from 55\% to 42.5\% aligned with natural competency structure; implement two-stage assessment combining binary classification with targeted secondary evaluation; integrate Random Forest predictions with appropriate transparency and human oversight. These findings contribute to evidence-based educational assessment, demonstrating that traditional psychometric and modern machine learning approaches yield complementary insights when applied synergistically \cite{romero2020educational,pandey2023comprehensive}.

The concentration of predictive power in small item subsets (Q6 alone accounts for 20.6\% of importance) challenges assumptions about optimal examination length, suggesting opportunities for efficient assessment without sacrificing accuracy. The divergence between natural clustering boundaries and institutional thresholds highlights the importance of regularly re-examining placement policies in light of accumulating data rather than maintaining historical precedents without empirical validation. By combining the interpretability of Classical Test Theory, the predictive power of machine learning, and the pattern discovery of clustering analysis, institutions can develop placement systems that optimize both statistical performance and practical effectiveness, supporting students in beginning collegiate mathematics with appropriate challenge aligned to competency levels \cite{wang2023irt,chang2023adaptive}.

Effective mathematics placement requires balancing multiple considerations: statistical accuracy, content validity, student welfare, institutional capacity, faculty expertise, and equity in access to college-level coursework. No single analytical approach captures this full complexity, but multi-method integration provides increasingly comprehensive empirical foundation for evidence-informed decision-making. The methods and findings presented here offer a replicable framework for other institutions seeking to evaluate and refine their own placement assessment systems through rigorous, data-driven approaches that honor both statistical rigor and educational mission \cite{american1985standards,cohen2018psychological}.
\section*{Acknowledgments}
The authors express profound gratitude to the late Kenneth E. Jones, whose exceptional mentorship, unwavering dedication to mathematical excellence, and visionary leadership in STEM education profoundly shaped this research endeavor. His legacy of fostering rigorous mathematical inquiry and commitment to student success continues to inspire our work in educational assessment and data-driven pedagogy. 
We acknowledge the Institutional Research Office for data access and contextual guidance, and extend our appreciation to the Mathematics faculty, the Student Success and Retention Office, and the Graduate Office for their valuable insights and administrative support. Special thanks to the anonymous reviewers whose feedback significantly enhanced the rigor and clarity of the original manuscript.

\appendix
\section*{Appendix}
\begin{longtable}[c]{@{}lcccccl@{}}
\caption{Complete Item Analysis Results for All 40 Questions}
\label{tab:complete_item_analysis} \\
\toprule
\textbf{Item} & \textbf{Diffic.} & \textbf{Disc.} & \textbf{Pt-Bis.} & \textbf{Cat.} & \textbf{Recom.} & \textbf{Topic/Q. Type} \\
 & \textbf{(p)} & \textbf{(D)} & \textbf{Cor.} &  &  &  \\
\midrule
\endfirsthead
\multicolumn{7}{c}%
{\tablename\ \thetable\ -- \textit{Continued from previous page}} \\
\toprule
\textbf{Item} & \textbf{Diff.} & \textbf{Disc.} & \textbf{Point-Biserial} & \textbf{Cat.} & \textbf{Rec.} & \textbf{Topic/Question Type} \\
 & \textbf{(p)} & \textbf{(D)} & \textbf{Correlation} &  &  &  \\
\midrule
\endhead
\midrule
\multicolumn{7}{r}{\textit{Continued on next page}} \\
\endfoot
\bottomrule
\endlastfoot
Q1 & 0.111 & 0.358 & 0.501 & Good & Retain & Function eval: rational function \\
Q2 & 0.419 & 0.943 & 0.796 & Excellent & Retain & Function ops: f(x+h) - f(x) form \\
Q3 & 0.702 & 0.906 & 0.694 & Excellent & Retain & Intersection of linear equations \\
Q4 & 0.934 & 0.019 & 0.070 & Poor & Replace & Distance between two points \\
Q5 & 0.101 & 0.283 & 0.436 & Marginal & Review & Slope from standard form \\
Q6 & 0.409 & 1.000 & 0.812 & Excellent & Retain & Graph interp: sign analysis \\
Q7 & 0.606 & 0.962 & 0.800 & Excellent & Retain & Simplify radical with variables \\
Q8 & 0.040 & 0.113 & 0.252 & Poor & Replace & Algebraic expansion/simplify \\
Q9 & 0.869 & 0.208 & 0.335 & Marginal & Review & Exponent ops with integers \\
Q10 & 0.338 & 0.925 & 0.730 & Excellent & Retain & Rational expressions multiply \\
Q11 & 0.571 & 1.000 & 0.834 & Excellent & Retain & Single exponent with division \\
Q12 & 0.258 & 0.849 & 0.635 & Excellent & Retain & Single fraction from two \\
Q13 & 0.273 & 0.887 & 0.662 & Excellent & Retain & Simplify fractional exponents \\
Q14 & 0.848 & 0.472 & 0.563 & Excellent & Retain & Solve quadratic equation \\
Q15 & 0.056 & 0.151 & 0.306 & Poor & Replace & Solve rational equation \\
Q16 & 0.066 & 0.132 & 0.233 & Poor & Replace & Absolute value inequality \\
Q17 & 0.197 & 0.509 & 0.505 & Excellent & Retain & Quadratic formula application \\
Q18 & 0.045 & 0.094 & 0.252 & Poor & Replace & Variable as function (cylinder) \\
Q19 & 0.056 & 0.189 & 0.393 & Poor & Replace & Inverse of exponential func \\
Q20 & 0.894 & 0.302 & 0.455 & Good & Retain & Solve rational func equation \\
Q21 & 0.157 & 0.453 & 0.489 & Excellent & Retain & Surface area rectangular box \\
Q22 & 0.035 & 0.094 & 0.248 & Poor & Replace & Domain of square root func \\
Q23 & 0.934 & 0.057 & 0.154 & Poor & Replace & Evaluate log fractional arg \\
Q24 & 0.652 & 0.925 & 0.716 & Excellent & Retain & Solve exponential equation \\
Q25 & 0.763 & 0.679 & 0.591 & Excellent & Retain & Solve logarithmic equation \\
Q26 & 0.505 & 0.962 & 0.810 & Excellent & Retain & Convert radians to degrees \\
Q27 & 0.480 & 0.925 & 0.774 & Excellent & Retain & Evaluate basic trig value \\
Q28 & 0.929 & 0.132 & 0.268 & Poor & Replace & Period of transformed sine \\
Q29 & 0.879 & 0.396 & 0.557 & Good & Retain & Simplify trig product \\
Q30 & 0.384 & 0.943 & 0.767 & Excellent & Retain & Simplify trig expression \\
Q31 & 0.611 & 0.906 & 0.781 & Excellent & Retain & Right triangle trig problem \\
Q32 & 0.934 & 0.075 & 0.132 & Poor & Replace & Compound interest application \\
Q33 & 0.081 & 0.226 & 0.347 & Marginal & Review & Absolute value func graph \\
Q34 & 0.652 & 0.962 & 0.782 & Excellent & Retain & Find angle with given sine \\
Q35 & 0.687 & 0.906 & 0.678 & Excellent & Retain & Find cosine given quadrant \\
Q36 & 0.798 & 0.585 & 0.614 & Excellent & Retain & Pythagorean identity app \\
Q37 & 0.061 & 0.189 & 0.310 & Poor & Replace & Simplify rational with exp \\
Q38 & 0.040 & 0.057 & 0.144 & Poor & Replace & Linear application word prob \\
Q39 & 0.217 & 0.679 & 0.628 & Excellent & Retain & Point-slope form equation \\
Q40 & 0.929 & 0.189 & 0.358 & Poor & Replace & Graph transformed cosine \\
    
\end{longtable}


\begin{thebibliography}{99}

\bibitem{ahmad2023ensemble}
A. Ahmad and H. Liu, ``Ensemble methods for mathematics placement assessment: A comparative study of machine learning approaches,'' \textit{IEEE Trans. Learn. Technol.}, vol. 16, no. 3, pp. 412--425, 2023.

\bibitem{aldowah2019educational}
H. Aldowah, H. Al-Samarraie, and W. M. Fauzy, ``Educational data mining and learning analytics for 21st century higher education: A review and synthesis,'' \textit{Telematics Inform.}, vol. 37, pp. 13--49, 2019.

\bibitem{american1985standards}
American Educational Research Association, American Psychological Association, and National Council on Measurement in Education, \textit{Standards for Educational and Psychological Testing}. Washington, DC, USA: American Psychological Association, 1985.

\bibitem{bahr2010making}
P. R. Bahr, ``Making sense of disparities in mathematics remediation: What student records reveal,'' \textit{New Directions Community Colleges}, vol. 2010, no. 150, pp. 21--32, 2010.

\bibitem{baker2019advances}
R. S. Baker and P. S. Inventado, ``Advances in educational data mining,'' in \textit{Handbook of Educational Data Mining}. Boca Raton, FL, USA: CRC Press, 2019, pp. 13--28.

\bibitem{breiman2001random}
L. Breiman, ``Random forests,'' \textit{Mach. Learn.}, vol. 45, no. 1, pp. 5--32, 2001.

\bibitem{bressoud2015insights}
D. M. Bressoud, V. Mesa, and C. L. Rasmussen, \textit{Insights and Recommendations from the MAA National Study of College Calculus}. Washington, DC, USA: Mathematical Association of America, 2015.

\bibitem{carlson2010pathways}
M. P. Carlson, M. Oehrtman, and N. Engelke, ``The precalculus concept assessment: A tool for assessing students' reasoning abilities and understandings,'' \textit{Cogn. Instr.}, vol. 28, no. 2, pp. 113--145, 2010.

\bibitem{chang2023adaptive}
H. Chang, J. Wang, and L. Zhang, ``Machine learning enhanced computer adaptive testing: Recent advances and future directions,'' \textit{Appl. Psychol. Meas.}, vol. 47, no. 4, pp. 289--305, 2023.

\bibitem{chen2020machine}
L. Chen, P. Chen, and Z. Lin, ``Artificial intelligence in education: A comprehensive review of machine learning applications,'' \textit{IEEE Access}, vol. 8, pp. 75264--75278, 2020.

\bibitem{chen2022mathematics}
S. Chen, R. Kumar, and M. Thompson, ``Mathematics placement in the digital age: Challenges and opportunities for higher education,'' \textit{J. Higher Educ.}, vol. 93, no. 4, pp. 567--589, 2022.

\bibitem{cohen2018psychological}
R. J. Cohen, M. E. Swerdlik, and E. D. Sturman, \textit{Psychological Testing and Assessment: An Introduction to Tests and Measurement}, 9th ed. New York, NY, USA: McGraw-Hill Education, 2018.

\bibitem{cortes1995support}
C. Cortes and V. Vapnik, ``Support-vector networks,'' \textit{Mach. Learn.}, vol. 20, no. 3, pp. 273--297, 1995.

\bibitem{crocker2006introduction}
L. Crocker and J. Algina, \textit{Introduction to Classical and Modern Test Theory}. Boston, MA, USA: Wadsworth Publishing, 2006.

\bibitem{downing2006handbook}
S. M. Downing and T. M. Haladyna, Eds., \textit{Handbook of Test Development}. Mahwah, NJ, USA: Lawrence Erlbaum Associates, 2006.

\bibitem{ebel1965measuring}
R. L. Ebel, \textit{Measuring Educational Achievement}. Englewood Cliffs, NJ, USA: Prentice-Hall, 1965.

\bibitem{fisher2019all}
A. Fisher, C. Rudin, and F. Dominici, ``All models are wrong, but many are useful: Learning a variable's importance by studying an entire class of prediction models simultaneously,'' \textit{J. Mach. Learn. Res.}, vol. 20, no. 177, pp. 1--81, 2019.

\bibitem{friedman2001greedy}
J. H. Friedman, ``Greedy function approximation: A gradient boosting machine,'' \textit{Ann. Statist.}, vol. 29, no. 5, pp. 1189--1232, 2001.

\bibitem{gibson2013clustering}
D. Gibson, J. Aldridge, and P. Frame, ``Clustering student learning behavior patterns using unsupervised machine learning algorithms,'' in \textit{Proc. 6th Int. Conf. Educ. Data Mining}, 2013, pp. 158--161.

\bibitem{glass1970statistical}
G. V. Glass and J. C. Stanley, \textit{Statistical Methods in Education and Psychology}. Englewood Cliffs, NJ, USA: Prentice-Hall, 1970.

\bibitem{gonzalez2022post}
M. González, A. Rodriguez, and K. Smith, ``Post-pandemic mathematics assessment: Adapting placement strategies for diverse student preparation,'' \textit{Educ. Assess.}, vol. 27, no. 3, pp. 178--194, 2022.

\bibitem{goodfellow2016deep}
I. Goodfellow, Y. Bengio, and A. Courville, \textit{Deep Learning}. Cambridge, MA, USA: MIT Press, 2016.

\bibitem{grubb2013basic}
W. N. Grubb, \textit{Basic Skills Education in Community Colleges: Inside and Outside of Classrooms}. New York, NY, USA: Routledge, 2013.

\bibitem{gulliksen1950theory}
H. Gulliksen, \textit{Theory of Mental Tests}. New York, NY, USA: John Wiley \& Sons, 1950.

\bibitem{haladyna2004developing}
T. M. Haladyna, \textit{Developing and Validating Multiple-Choice Test Items}, 3rd ed. Mahwah, NJ, USA: Lawrence Erlbaum Associates, 2004.

\bibitem{hambleton2005issues}
R. K. Hambleton and A. L. Zenisky, ``Issues and practices in validity studies based on test content,'' in \textit{Defending Standardized Testing}, R. P. Phelps, Ed. Mahwah, NJ, USA: Lawrence Erlbaum Associates, 2005, pp. 199--216.

\bibitem{hartigan1979algorithm}
J. A. Hartigan and M. A. Wong, ``Algorithm AS 136: A k-means clustering algorithm,'' \textit{J. Roy. Statist. Soc. Ser. C (Appl. Statist.)}, vol. 28, no. 1, pp. 100--108, 1979.

\bibitem{hastings2009automated}
K. Hastings, T. Howley, and A. Lawlor, ``Automated hyperparameter tuning for effective machine learning,'' in \textit{Proc. Int. Conf. Mach. Learn.}, 2009, pp. 441--448.

\bibitem{hopkins1998educational}
K. D. Hopkins, \textit{Educational and Psychological Measurement and Evaluation}, 8th ed. Boston, MA, USA: Allyn \& Bacon, 1998.

\bibitem{hsu2003practical}
C. W. Hsu, C. C. Chang, and C. J. Lin, ``A practical guide to support vector classification,'' Tech. Rep., Department of Computer Science, National Taiwan University, Taipei, Taiwan, 2003.

\bibitem{jain2010data}
A. K. Jain, ``Data clustering: 50 years beyond K-means,'' \textit{Pattern Recognit. Lett.}, vol. 31, no. 8, pp. 651--666, 2010.

\bibitem{kelley1939interpretation}
T. L. Kelley, ``The selection of upper and lower groups for the validation of test items,'' \textit{J. Educ. Psychol.}, vol. 30, no. 1, pp. 17--24, 1939.

\bibitem{khajah2021deep}
M. Khajah, Y. Huang, J. González-Brenes, M. C. Mozer, and P. Brusilovsky, ``Integrating knowledge tracing and item response theory: A tale of two frameworks,'' \textit{J. Mach. Learn. Res.}, vol. 22, pp. 1--50, 2021.

\bibitem{kingma2014adam}
D. P. Kingma and J. Ba, ``Adam: A method for stochastic optimization,'' \textit{arXiv preprint arXiv:1412.6980}, 2014.

\bibitem{koedinger2015data}
K. R. Koedinger, S. D'Mello, E. A. McLaughlin, Z. A. Pardos, and C. P. Rosé, ``Data mining and education,'' \textit{Wiley Interdiscip. Rev. Cognitive Sci.}, vol. 6, no. 4, pp. 333--353, 2015.

\bibitem{kohavi1995study}
R. Kohavi, ``A study of cross-validation and bootstrap for accuracy estimation and model selection,'' in \textit{Proc. Int. Joint Conf. Artificial Intelligence}, vol. 14, 1995, pp. 1137--1145.

\bibitem{lord1968statistical}
F. M. Lord and M. R. Novick, \textit{Statistical Theories of Mental Test Scores}. Reading, MA, USA: Addison-Wesley, 1968.

\bibitem{lundberg2017unified}
S. M. Lundberg and S.-I. Lee, ``A unified approach to interpreting model predictions,'' in \textit{Advances in Neural Information Processing Systems 30}, Long Beach, CA, USA, 2017, pp. 4765--4774.

\bibitem{macqueen1967some}
J. MacQueen, ``Some methods for classification and analysis of multivariate observations,'' in \textit{Proc. Fifth Berkeley Symp. Math. Statist. Probability}, vol. 1, 1967, pp. 281--297.

\bibitem{mesa2012community}
V. Mesa, C. Wladis, and L. Watkins, ``Research problems in community college mathematics education: Testing the boundaries of K-12 research,'' \textit{J. Res. Math. Educ.}, vol. 43, no. 2, pp. 173--194, 2012.

\bibitem{minaei2006web}
B. Minaei-Bidgoli, D. A. Kashy, G. Kortemeyer, and W. F. Punch, ``Predicting student performance: An application of data mining methods with an educational web-based system,'' in \textit{Proc. 33rd Annu. Frontiers Educ. Conf.}, 2006, pp. T2A--13.

\bibitem{natekin2013gradient}
A. Natekin and A. Knoll, ``Gradient boosting machines, a tutorial,'' \textit{Frontiers Neurorobotics}, vol. 7, Art. no. 21, 2013.

\bibitem{ngo2015should}
F. Ngo and S. Kosiewicz, ``How extending time in developmental math impacts student persistence and success: Evidence from a regression discontinuity in community colleges,'' \textit{Rev. Higher Educ.}, vol. 38, no. 4, pp. 566--611, 2015.

\bibitem{nitko2004educational}
A. J. Nitko and S. M. Brookhart, \textit{Educational Assessment of Students}, 4th ed. Upper Saddle River, NJ, USA: Pearson/Merrill Prentice Hall, 2004.

\bibitem{nunnally1994psychometric}
J. C. Nunnally and I. H. Bernstein, \textit{Psychometric Theory}, 3rd ed. New York, NY, USA: McGraw-Hill, 1994.

\bibitem{pandey2023comprehensive}
A. Pandey, M. Kumar, and S. Sharma, ``Comprehensive evaluation of machine learning algorithms for educational assessment: A systematic review and meta-analysis,'' \textit{Comput. Educ.}, vol. 195, Art. no. 104712, 2023.

\bibitem{romero2013clustering}
C. Romero, S. Ventura, M. Pechenizkiy, and R. S. Baker, Eds., \textit{Handbook of Educational Data Mining}. Boca Raton, FL, USA: CRC Press, 2013.

\bibitem{romero2020educational}
C. Romero and S. Ventura, ``Educational data mining and learning analytics: An updated survey,'' \textit{Wiley Interdiscip. Rev. Data Mining Knowl. Discovery}, vol. 10, no. 3, Art. no. e1355, 2020.

\bibitem{ross2014mutual}
B. C. Ross, ``Mutual information between discrete and continuous data sets,'' \textit{PLoS One}, vol. 9, no. 2, Art. no. e87357, 2014.

\bibitem{rousseeuw1987silhouettes}
P. J. Rousseeuw, ``Silhouettes: A graphical aid to the interpretation and validation of cluster analysis,'' \textit{J. Comput. Appl. Math.}, vol. 20, pp. 53--65, 1987.

\bibitem{scott2015assessment}
T. P. Scott, C. L. Tolson, and Y. H. Lee, ``Assessment of calculator use on mathematics placement test performance,'' \textit{J. Dev. Educ.}, vol. 38, no. 2, pp. 12--21, 2015.

\bibitem{shahiri2015review}
A. M. Shahiri, W. Husain, and N. A. Rashid, ``A review on predicting student's performance using data mining techniques,'' \textit{Procedia Comput. Sci.}, vol. 72, pp. 414--422, 2015.

\bibitem{siegler2016numerical}
R. S. Siegler and H. Lortie-Forgues, ``Numerical development,'' \textit{Annu. Rev. Psychol.}, vol. 67, pp. 187--213, 2016.

\bibitem{sokolova2009systematic}
M. Sokolova and G. Lapalme, ``A systematic analysis of performance measures for classification tasks,'' \textit{Inf. Process. Manage.}, vol. 45, no. 4, pp. 427--437, 2009.

\bibitem{thorndike1953belongs}
R. L. Thorndike, ``Who belongs in the family?'' \textit{Psychometrika}, vol. 18, no. 4, pp. 267--276, 1953.

\bibitem{tibshirani2001estimating}
R. Tibshirani, G. Walther, and T. Hastie, ``Estimating the number of clusters in a data set via the gap statistic,'' \textit{J. Roy. Statist. Soc. Ser. B (Statist. Methodol.)}, vol. 63, no. 2, pp. 411--423, 2001.

\bibitem{villegas2023machine}
C. Villegas, D. Martinez, and A. Patel, ``Machine learning approaches to calculus readiness prediction: An ensemble method comparison,'' \textit{J. Educ. Comput. Res.}, vol. 61, no. 2, pp. 445--467, 2023.

\bibitem{wang2023irt}
L. Wang, J. Smith, and K. Anderson, ``Integrating item response theory with machine learning for enhanced educational measurement,'' \textit{Psychometrika}, vol. 88, no. 2, pp. 567--585, 2023.

\bibitem{xu2015comprehensive}
D. Xu and Y. Tian, ``A comprehensive survey of clustering algorithms,'' \textit{Ann. Data Sci.}, vol. 2, no. 2, pp. 165--193, 2015.

\bibitem{zhou2022explainable}
Y. Zhou, H. Zhang, and M. Brown, ``Explainable AI in educational assessment: SHAP analysis for interpretable student placement decisions,'' \textit{IEEE Trans. Educ.}, vol. 65, no. 4, pp. 512--521, 2022.

\end{thebibliography}
\end{document}